\documentclass[runningheads]{llncs}

\usepackage{eccv}

\usepackage{eccvabbrv}

\usepackage{graphicx}
\usepackage{booktabs}
\usepackage{subcaption}

\usepackage{wrapfig}
\usepackage[accsupp]{axessibility}  

\usepackage{hyperref}

\usepackage{orcidlink}

\begin{document}

\title{DiffusionPen: Towards Controlling the Style of Handwritten Text Generation} 

\titlerunning{DiffusionPen: Towards Controlling the Style of Handwritten Text Generation}

\author{Konstantina Nikolaidou\inst{1}\orcidlink{0000-0002-9332-3188} \and George Retsinas\inst{2}\orcidlink{0000-0001-6734-3575} \and Giorgos Sfikas\inst{3}\orcidlink{0000-0002-7305-2886}
 \and Marcus Liwicki\inst{1}\orcidlink{0000-0003-4029-6574}}

\authorrunning{K.~Nikolaidou et al.}


\institute{
Luleå University of Technology, Sweden \\
\email{firstname.lastname@ltu.se}\\
\and
National Technical University of Athens, Greece\\
\email{gretsinas@central.ntua.gr}\\
\and
University of West Attica, Greece\\
\email{gsfikas@uniwa.gr}
}

\maketitle

\begin{abstract}

Handwritten Text Generation (HTG) conditioned on text and style is a challenging task due to the variability of inter-user characteristics and the unlimited combinations of characters that form new words unseen during training.
Diffusion Models have recently shown promising results in HTG but still remain under-explored.
We present DiffusionPen (DiffPen), a 5-shot style handwritten text generation approach based on Latent Diffusion Models. 
By utilizing a hybrid style extractor that combines metric learning and classification, our approach manages to capture both textual and stylistic characteristics of seen and unseen words and styles, generating realistic handwritten samples.
Moreover, we explore several variation strategies of the data with multi-style mixtures and noisy embeddings, enhancing the robustness and diversity of the generated data. 
Extensive experiments using IAM offline handwriting database show that our method outperforms existing methods qualitatively and quantitatively, and its additional generated data can improve the performance of Handwriting Text Recognition (HTR) systems. 
The code is available at: \url{https://github.com/koninik/DiffusionPen}.

\end{abstract}
\section{Introduction}
\label{sec:intro}

Handwritten Text Generation (HTG) or Styled HTG is a challenging task recently gaining increased attention.
The challenge lies in preserving the readability of specific textual content while capturing the unique characteristics of a writer.
The ability to automatically generate text that resembles a specific writing style could enhance personalization in digital design or potentially assist people facing writing challenges.
Furthermore, more relevant to this work, it enables the augmentation of datasets to train efficient text recognition systems.

\begin{wrapfigure}{l}{7.1cm}
    \centering
\includegraphics[width=0.6\textwidth]{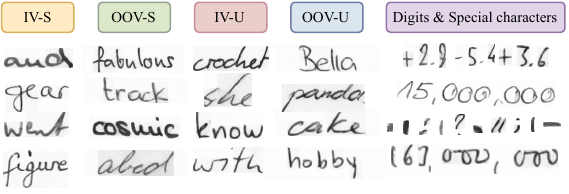}
    \caption{Qualitative results generated using our method for four cases: In-Vocabulary words and Seen style (IV-S), In-Vocabulary words and Unseen style (IV-U), Out-of-Vocabulary words and Seen style (OOV-S), Out-of-Vocabulary words and Unseen style (OOV-U), as well as digits and punctuations.}
    \vspace{-.5cm}
    \label{fig:vocabulary_styles}
\end{wrapfigure}

Generative Adversarial Networks (GANs) have been the predominant method for offline HTG~\cite{kang2020ganwriting,Fogel2020ScrabbleGANSV,alonso2019adversarial, Davis2020TextAS, mattick2021smartpatch}.
In terms of network architecture, Transformer-based solutions~\cite{vaswani2017attention,Bhunia_2021_ICCV,Pippi2023HandwrittenTG}
are invariably employed, following the trends set in other fields.
Among these standard methods, Denoising Diffusion Probabilistic Models (DDPM)~\cite{ho2020denoising} have recently emerged as a compelling alternative for HTG, offering a new paradigm distinct from traditional GANs and showcasing impressive results~\cite{zhu2023conditional, nikolaidou2023wordstylist}.
A common approach in GAN-based methods concerning the treatment of handwriting style is to incorporate a writer recognizer to classify the generated samples during training in order to force the generator to learn how to generate a specific handwriting style.
However, adversarial training is known to suffer from limited diversity in the generated samples and presents instabilities in the training process \cite[\S 15.1.4]{prince2023understanding}.
The same approach is not straightforward when using DDPM since the objective during training is to model the noise.
Thus, the style space must be modeled more carefully.

DDPMs are a class of hierarchical Variational Autoencoders (VAEs) ~\cite{SohlDickstein2015DeepUL,ho2020denoising,kingma2021variational} that have recently garnered
considerable interest within the representation learning and vision communities.
Among an assortment of impressive results on a range of tasks, 
they have notably dominated the field of text-to-image generation by creating high-quality images given a text prompt~\cite{rombach2022high,nichol2021glide,ramesh2022hierarchical,saharia2022photorealistic}.
Their success relies on the efficiency of the model itself and the use of pre-training techniques of large-scale image-text pairs~\cite{Radford2021LearningTV}. 
While numerous diffusion-based systems demonstrate high-quality results in generating images given a text description~\cite{rombach2022high,nichol2021glide,ramesh2022hierarchical,saharia2022photorealistic, balaji2022ediffi}, fewer works focus on generating 
readable scene-text images~\cite{chen2023textdiffuser, zhu2023conditional,zhang2023brush} or fonts~\cite{he2022diff, yang2023fontdiffuser} and, related to this work, generating handwriting~\cite{zhu2023conditional, nikolaidou2023wordstylist,gui2023zero}. 

In this work, we present a latent diffusion model that generates handwritten text images conditioned on a text prompt and a limited set of style samples in a few-shot scheme.
As can be seen in~\cref{fig:vocabulary_styles}, our proposed method manages to generate realistic samples of seen and unseen styles as well as In-Vocabulary (IV) and Out-of-Vocabulary (OOV) words.
Most importantly, we deal successfully with the problem of limited diversity when sampling from the posterior and attempt to manipulate the output samples through various strategies.
An effect relating conditional weighting and stereotypical sampling has been recently discussed in the context of diffusion-based modeling \cite[\S 18.6.3]{prince2023understanding}. 
To the best of our knowledge, this is one of the first works incorporating the few-shot style scheme in 
diffusion-based methods for HTG.
We show that the resulting model leads to handwriting samples of simultaneously high diversity and high quality while conditioned on textual and style information.

\noindent
\textbf{Contributions.}
We propose \textit{DiffusionPen} (\textit{DiffPen}), a styled handwritten text generation method based on latent diffusion models.
The method comprises a latent denoising autoencoder that performs the denoising diffusion process as the 
main network, and two auxiliary pre-trained encoders to create the style and textual conditions.
The style encoder is based on the combination of classification and metric-learning training, which creates a continuous space for the style embeddings, providing more diversity to the generation process.
The style condition is introduced in the main network in a few-shot setting to represent the unique characteristics of each writer from a limited set of $k=5$ samples.

Our method is able to imitate the style of a writer given specific text content and five images from the specific writer.
In particular, we show that we:
\begin{itemize}
    \item \emph{avoid posterior collapse}; given a text and style embedding, the proposed model is capable of producing highly diverse handwriting samples. 
    \item \emph{estimate a meaningful style space}; points in the style space invariably correspond to realistic, unseen styles.
    \item \emph{outperform numerically the current state of the art by a significant margin}.
    \item \emph{control style generation via style interpolation, style mixture \& noise bias}.
\end{itemize}

We evaluate our proposed method by presenting both qualitative and quantitative results.
Through qualitative results, we show that our method is able to generate  IV and OOV words of both seen and unseen writer styles.
We quantify generated data quality by computing commonly used metrics and comparing versus other SotA methods.
Furthermore, we quantify style quality by examining whether a writer recognizer trained on real data can recognize the writer class of the generated data. 
To quantify the diversity of the generated data, we conduct extensive experiments using an ``auxiliary'' HTR task; in particular, we use our model to imitate the real dataset and proceed to explore the variation present in the generated data by measuring the extent of improvement over HTR performance after using diffusion-generated data in the training process.
Moreover, we present different sampling strategies that incorporate noise bias, style interpolation, and style mixture that showcase how style can be controlled and give extra variation to the generation.
Finally, we present limitations with practical solutions, leading to future work perspectives, and discuss ethical considerations.

\section{Related Work}
\label{sec:related_work}


\textbf{Handwritten Text Generation.}
The steady progress in the expressiveness and sophistication of generative modeling
has enabled HTG, especially after the advent of adversarial modeling.
Most works focusing on offline HTG rely on GAN-based approaches.
Alonso \etal~\cite{alonso2019adversarial} present a GAN-based approach that takes as input a sequential text embedding encoded by an LSTM Recurrent Neural Network and further deploys an auxiliary network that uses CTC loss to recognize the generated text.  
Similarly, ScrabbleGAN~\cite{Fogel2020ScrabbleGANSV} uses a text recognizer to help improve the quality of the generated text and character filters, showing variability in style and stroke width.
Both approaches focus mostly on conditioning on the text content.
On the contrary, Davis \etal~\cite{Davis2020TextAS} present a method that conditions on both text and style to generate realistic handwritten lines of arbitrary length by predicting the space required between text.
Likewise, GANwriting~\cite{kang2020ganwriting} is a GAN-based system conditioned on text and few-shot stylistic samples and is trained in an adversarial manner with additional help from a text recognizer and a writer classification network.
The method manages to generate realistic handwritten images of in-vocabulary and out-of-vocabulary words of seen and unseen writer styles.
The work is further extended in~\cite{kang2021content} to also work for whole sentences.
Although GANwriting generates understandable and stylistic samples there are several artifacts present in the generated data.  
An approach based on GANwriting named SmartPatch was introduced in ~\cite{mattick2021smartpatch} to tackle these artifacts. 

Also based on a GAN framework and adversarial training, ~\cite{Bhunia_2021_ICCV} and~\cite{Pippi2023HandwrittenTG} have combined the encoder-decoder nature of Transformers with a few-shot style encoding to generate handwritten text.
Further following the image synthesis trends, the works presented in~\cite{nikolaidou2023wordstylist} and~\cite{zhu2023conditional} have introduced the application of Diffusion Models to synthesize understandable handwritten text.
These systems have the ability to generate high quality text conditioning on a writer style and a text content, however they are limited in the way they represent and handle unseen styles.
In this work, we address the limitations of the aforementioned approaches~\cite{nikolaidou2023wordstylist,zhu2023conditional} and propose a Diffusion-based generative model that can produce unseen writing style samples by deploying pre-trained writer classifiers in a few-shot setting.

\noindent
\textbf{Few-Shot Conditional Diffusion Models.}
Few-shot Conditional Generation is the task of generating new samples of a specific class or object by conditioning on a few samples instead of a class embedding.
This further enables the generation of unseen classes.
Few-Shot Diffusion Models~\cite{Giannone2022FewShotDM} condition the generation on a small set of image patches using a Vision Transformer (ViT).
D2C~\cite{Sinha2021D2CDM} is a conditional few-shot Latent Diffusion Model that utilizes contrastive self-supervision to learn the latent space.
The existing work indicates that there is plenty of room to explore conditioning diffusion models in few-shot schemes.

\section{Proposed Method}
\label{sec:method}

The problem formulation of this work can be described as follows.
Given $k=5$ samples written by a writer $w\in W$ and a word $t$ comprising $i$ characters, our goal is to generate new images $Y_w^t$ that depict the text content in $t$ and the style of writer $w$.
This task can be cast in terms of a conditional generative model,
where we need to learn a distribution of handwriting samples $q(\cdot)$.
Sampling over the distribution (conditioned on $w,t$) will produce the desired new images $Y_w^t$.

Prior work on HTG, using similar considerations with GAN-based approaches \cite{Bhunia_2021_ICCV,Pippi2023HandwrittenTG,kang2020ganwriting} or diffusion modeling \cite{nikolaidou2023wordstylist,zhu2023conditional} is hindered by two correlated issues.
First, the style space is inadequately modeled in the sense that points in the sample space are not guaranteed to correspond to a meaningful style.
This is particularly visible in the results of~\cref{sec:experiments}, where some of the compared methods achieve very high CER and WER scores when used to train an HTR system, indicating that they lack variation 
due to mode collapse.
Second, sampling from the posterior given style and content gives samples that are practically too close
to specific distribution modes.

We deploy a Conditional Latent Diffusion Model in combination with an existing text encoder and a feature extractor that operates in a few-shot scheme on the style samples.
The feature extractor is trained using a hybrid metric-learning and classification approach to obtain a more intuitive feature space for the writer-style representations.
In this manner, we constrain the learned style space to retain a sense of prescribed style distance. 

\begin{figure*}[t]
  \centering
  \begin{subfigure}{0.73\textwidth}
        \centering
        \includegraphics[width=\linewidth]{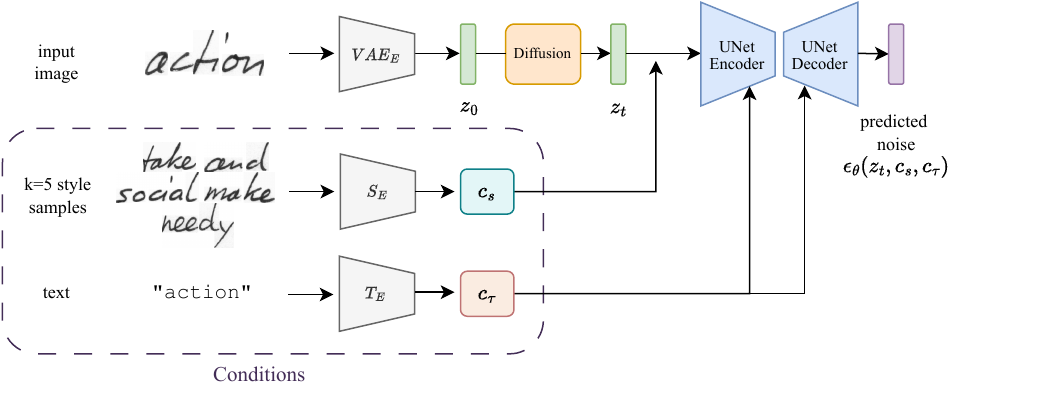}
        \caption{DiffusionPen Training Pipeline.}
        \label{fig:top}
    \end{subfigure}
    \begin{subfigure}{0.73\textwidth}
        \centering
        \includegraphics[width=\linewidth]{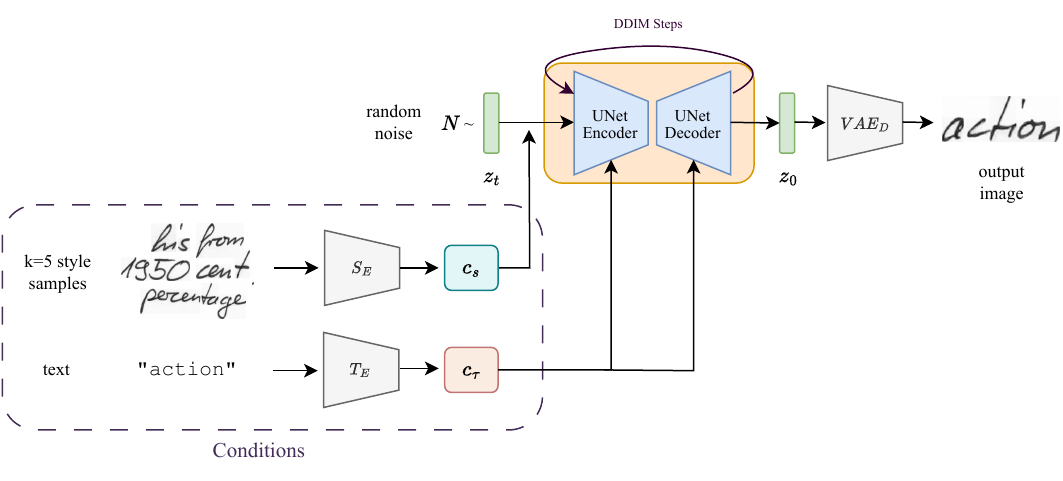}
        \caption{DiffusionPen Sampling Pipeline.}
        \label{fig:bottom}
    \end{subfigure}
   
   \caption{Overview of \emph{DiffusionPen}. \emph{DiffusionPen} comprises the conditional generator UNet Encoder-Decoder, having a Text Encoder $T_E$, a Style Encoder $S_E$, and a \emph{VAE\textsubscript{E}} encoder during training (\ref{fig:top}) and \emph{VAE\textsubscript{D}} decoder during sampling (\ref{fig:bottom}).}
   \label{fig:model}
\end{figure*}

\noindent
\textbf{Style Encoder.}
Given a batch of images, the goal of the style encoder $S_E$ is to extract meaningful feature representations that encapsulate the writer characteristics of each image to ultimately be used in a few-shot learning setting and condition the diffusion model training. 
To this end, we utilize a MobileNetV2 backbone~\cite{Sandler2018MobileNetV2IR} as the style encoder $S_E$ due to its high performance and lightweight design, and we combine a classification and metric learning approach during training.
A small ablation on the choice of the backbone is presented in the supplementary material.

\begin{wrapfigure}{r}{5.5cm}
\vspace{-.5cm}
    \centering
    \includegraphics[width=0.5\textwidth]{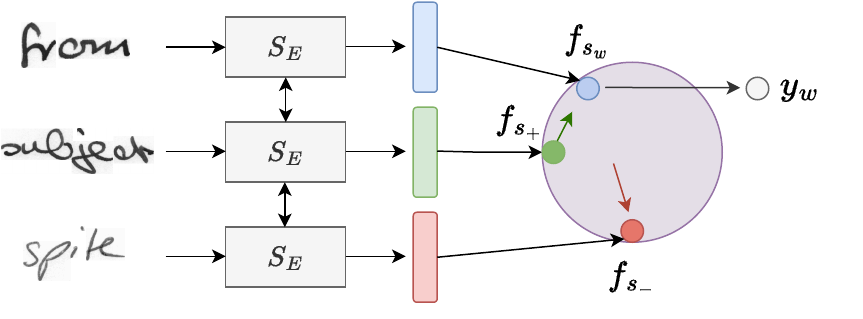}
    \caption{
    A graphical representation of the hybrid style encoder $S_E$ training. 
    The style encoder creates the feature representations of the anchor sample $f_{s_w}$, positive sample $f_{s_+}$, and negative sample $f_{s_-}$. The metric learning training part pushes the positive features closer to the anchor and the negative features further away. The model uses the class prediction $y_w$ of the anchor for the classification optimization.}
    \vspace{-.5cm}
    \label{fig:style_encoder}
\end{wrapfigure}

Given a sample image $s_w$, used as an anchor, the model learns its stylistic characteristics from a random positive sample $s_+$ from the same writer and a random negative sample $s_-$ from a different writer.
The model learns the similarity between the samples using a triplet loss $\mathcal{L}_{triplet}$ formulated as:
\(\mathcal{L}_{triplet}(s_w,s_+,s_-) = \max\left( 0, \delta_{+} - \delta_{-} + \alpha \right)\), where $\delta_\pm = \| f_{s_w} - f_{s_\pm} \|_p$ and $\alpha$ is a margin.
Furthermore, the feature representation of the anchor sample $f_{s_w}$ is passed through a classification layer to predict its writer class. 
The classification part is optimized using Cross-Entropy as the classification loss $\mathcal{L}_{class}$. 
The model is trained using the combination of the two parts, formulating the loss as: \(\mathcal{L}_{comb}=\mathcal{L}_{class}(f_{s_w},w) + \mathcal{L}_{triplet}(s_w,s_+,s_-)\).
A graphical representation of the style encoder hybrid training is presented in~\cref{fig:style_encoder}.
This hybrid approach provides a feature space that keeps the different classes well-separated and robustness in intra-class variation.
More details about the training of the style encoder are presented in~\cref{sub:training}.

Given the pre-trained style encoder, the style condition $c_s$ that is fed into the diffusion model is created as follows. 
For every image in the training set, we consider $k$ samples from the same writer class and pass them through $S_E$ to extract the feature representation of each style image $s_k$.
Then, we aggregate the extracted $d$-dimensional features of the style images by obtaining the mean of the $k$ feature embeddings $s_{emb}\in\mathbb{R}^{d}$.
Unlike most works that use 15 samples~\cite{kang2020ganwriting, mattick2021smartpatch, Pippi2023HandwrittenTG, Bhunia_2021_ICCV} to get the stylistic characteristics, we condition the writer style on $k=5$ style features.
Finally, the mean feature embedding is projected to the model dimension using a linear layer $F_{proj}$, giving the final style condition as \(c_s = F_{proj}(s_{emb})\in\mathbb{R}^{d_{model}}\).

\noindent
\textbf{Text Encoder.}
The text condition $c_\tau$ defines the textual content depicted in the images.
The condition is created by using \textsc{Canine}-C~\cite{Clark2021CaninePA} as the text encoder.
\textsc{Canine}-C is an encoder that operates directly on character sequences without using an explicit vocabulary.
This is particularly useful in the case of handwriting generation in order to generate OOV words.
First, the raw character input sequence $\tau$ is given to the \textsc{Canine} tokenizer to get a structured format of the words, giving a unique token to each character and padding every word to a maximum length for batch processing.
An encoded embedding $\tau_{emb}$ of the tokenized input is then created by the text encoder, and then, $F_{proj}$ is applied to the text embedding to obtain the text condition \(c_\tau = F_{proj}(\tau_{emb})\in\mathbb{R}^{d_{model}}\).
The text encoder and Conditional Latent Diffusion Model are trained using the objective described in the following paragraphs.


\noindent
\textbf{Conditional Latent Diffusion Model.}
Diffusion models can be understood as a special case of a Variational Autoencoder,
where the latent space is defined as a Markov chain consisting of random variables $z_1,\cdots,z_T$.
These variables have the same dimensionality as the initial sample $x_0$ and furthermore
the encoder is (usually) fixed, with Gaussian noise being added layer after layer of the Markov chain.
In diffusion modeling, we aim to learn the decoder or otherwise termed reverse or denoising phase,
to be understood as letting the network learn how to gradually remove noise from $z_T$ gradually back to the original sample space.

For the latent diffusion-based network, we utilize a UNet architecture~\cite{ronneberger2015u}, similar to WordStylist~\cite{nikolaidou2023wordstylist},
as the network that learns the noise distribution to be removed.
To reduce computational cost, we use a pretrained VAE encoder \cite{rombach2022high} to map the original image input of shape $W\times H$ into a 4-D latent representation $z\in\mathbb{R}^{4\times W/8 \times H/8}$ as input to the network that performs the diffusion and the denoising process.
In the forward diffusion process, a timestep $t\in{[0,T]}$ and Gaussian noise $\epsilon\in\mathbb{R}^{4\times W/8 \times H/8}$ are sampled to corrupt the initial latent representation $z_t$.
The network is trained using the denoising loss between the sampled Gaussian noise $\epsilon$ and the predicted noise $\epsilon_\theta$, as: \(L = {\|\epsilon - \epsilon_\theta(z_t, c_s, c_\tau)\|}^2_2\).
In the backward denoising process or sampling, given a style embedding and a text condition, the denoising autoencoder predicts and subtracts the present noise given the previous denoised sample. 
Finally, the predicted latent sample is given to the VAE decoder to create the final image.
~\cref{fig:model} presents the overall architecture of our method.
\section{Experiments}
\label{sec:experiments}

\subsection{Datasets, Training Setup, and Considered SotA Approaches}
\label{sub:training}

\textbf{Datasets.} IAM Offline Handwriting Database~\cite{marti2002iam} is one of the most commonly used datasets for handwriting recognition. 
It contains $\sim$115K isolated words and their transcriptions written by various writers.
Similar to~\cite{kang2020ganwriting}, we use 339 writers to train the style encoder and diffusion model, and we keep 160 for the experimental evaluation of the unseen style scenario and the HTR system.
Additionally, we use the GNHK dataset~\cite{lee2021gnhk}, which includes unconstrained camera-captured images of English handwritten text, and show qualitative results in the supplementary material.

\noindent
\textbf{Training Setup.} 
The training process occurs in two stages: the \emph{Style Encoder} and the \emph{Denoising Model} training.
The \emph{Style Encoder} is trained as described in~\cref{sec:method}, using IAM database.
The style extractor is trained for 20 epochs, with a batch size of 320, Adam~\cite{kingma2014adam} as the optimizer, and a learning rate of 0.001 that is reduced by a factor of 0.1 every 3 epochs and weight decay of 0.0001.
We used a random selection for the negative samples as the inherently varied and nuanced differences across writers reduce the chance of selecting an easier negative; hence, the randomly chosen examples are sufficiently challenging, and no convergence issues were observed.
All images are initially rescaled to a height of $64$ pixels, preserving the aspect ratio. 
If the width of an image after rescaling is less than $256$ pixels, padding is added to a fixed width of $256$ pixels. Otherwise, the image is resized in height and width until obtaining a width less than $256$ pixels and then padded in both height and width to a fixed size of $64 \times 256$.
The same image pre-processing is also used in the main model training.
The style encoder is trained independently from the diffusion model and is kept frozen during the diffusion training to create the stylistic feature condition.
For the \emph{Denoising Model} training, we use DDIM~\cite{song2020denoising} noise scheduler for the noise injection and sampling.
During training, the diffusion timesteps are set to 1K, while for sampling, the noise scheduler gives the flexibility to reduce the backward timesteps to 50.
AdamW~\cite{Loshchilov2017DecoupledWD} is the optimizer with a weight decay of 0.2 and a learning rate of 0.0001.
Every model is trained with a batch size of 320 for 1K epochs on a single A100 SXM GPU.

\noindent
\textbf{Considered SotA Approaches.}
For qualitative and quantitative comparison with the literature, we consider the GAN-based methods GANwriting~\cite{kang2020ganwriting} and SmartPatch~\cite{mattick2021smartpatch}, the Transformer-based VATr~\cite{Pippi2023HandwrittenTG}, and the Diffusion-based WordStylist~\cite{nikolaidou2023wordstylist}.
GANwriting, SmartPatch, and VATr are similar to our method in terms of few-shot style condition. 
However, these methods use 15 samples to create the style embedding, while we use only 5. 
Furthermore, these methods use an auxiliary writer identification network as the feature extractor that is trained dynamically for the task with the generator.
On the other hand, WordStylist is relevant to our method, as the main denoising diffusion process and network are similar, while the key difference is that WordStylist conditions on the style as a whole class embedding, which limits it to create only previously seen styles. 
\begin{figure}
    \vspace{-.5cm}
    \centering
\includegraphics[width=0.81\textwidth]{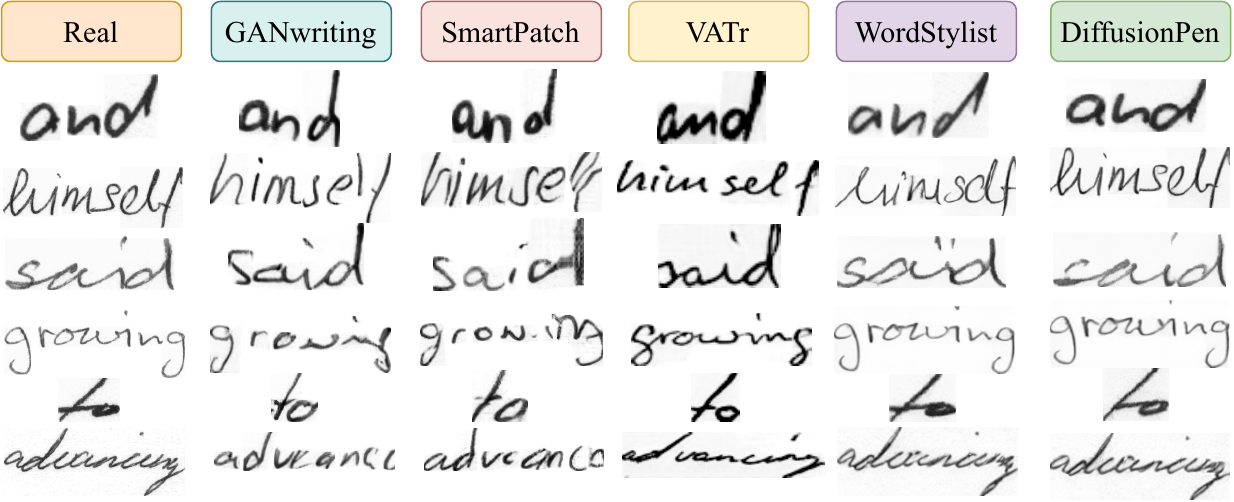}
    \caption{Visual comparison of images generated by the considered approaches and our proposed method (DiffusionPen).}
    \label{fig:comparison_sota}
\end{figure}

\vspace{-7.5mm}
\subsection{Quality Assessment}
\begin{figure}
    \centering
    \begin{subfigure}[b]{0.49\textwidth}
    \includegraphics[width=\linewidth]{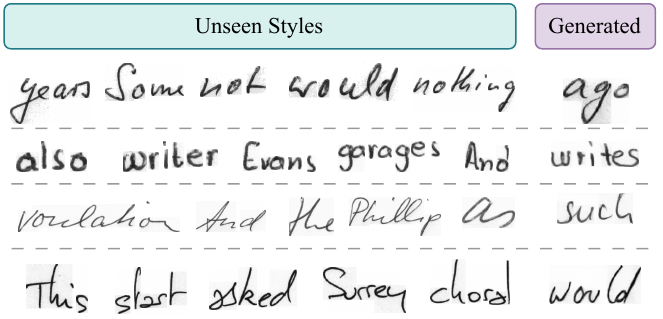}
    \caption{Unseen styles generation.}
    \label{fig:unseen_styles}
    \end{subfigure}
    \begin{subfigure}[b]{0.49\textwidth}
    \centering
    \includegraphics[width=\linewidth]{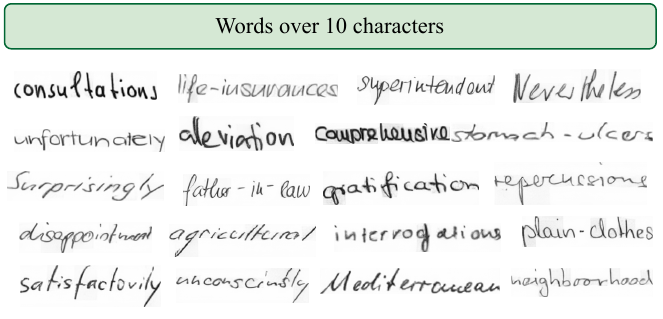}
    \caption{Words containing more than 10 characters.}
    \label{fig:10_chars}
    \end{subfigure}

    \caption{\textbf{(a)} Exemplar generated samples from Unseen Styles. On the left, we can see the 5 style samples used for the style condition, and on the right, the generated word. \textbf{(b)} Generated words of different styles comprised of more than 10 characters.}
    \label{fig:unseen_iv_oov_10chars}
\end{figure}

\begin{figure}
    \centering    \includegraphics[width=.99\linewidth]{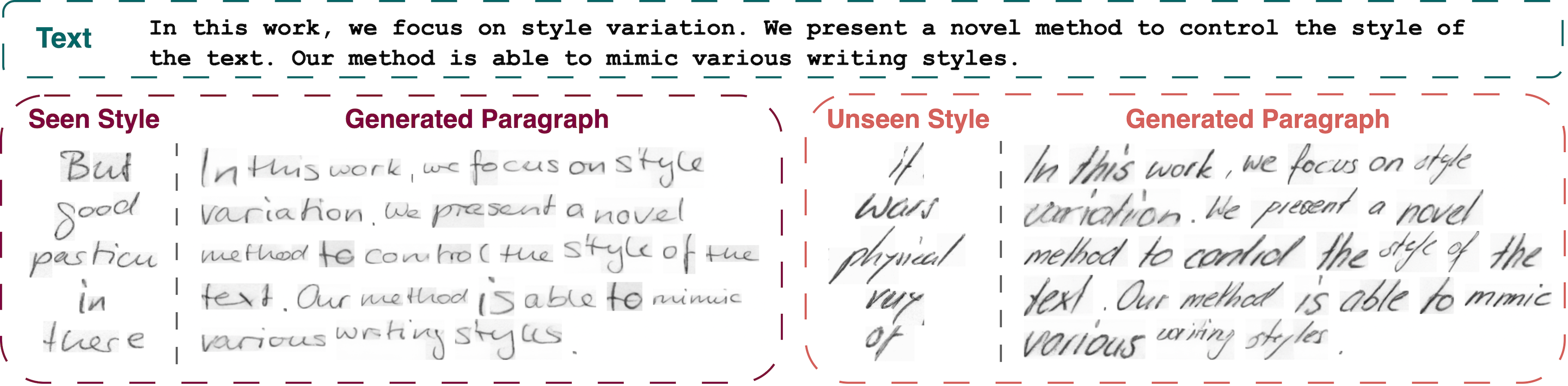}
    \caption{A small paragraph generated in a Seen and Unseen Style.}
    \label{fig:paragraphs}
\end{figure}

\cref{fig:vocabulary_styles} shows that our approach manages to generate samples of Seen (S) and Unseen (U) styles, In-Vocabulary (IV) and Out-of-Vocabulary (OOV) words, as well as digits and special characters.
Comparative visual results with the SotA methods are also presented in~\cref{fig:comparison_sota}.
Furthermore, examples of generated words containing more than 10 characters are presented in~\cref{fig:10_chars}, showcasing the ability of our model to create longer words.
Unseen styles are also presented in~\cref{fig:unseen_styles}.
Finally, we present small paragraphs generated using our method in~\cref{fig:paragraphs}. 
More visual examples of both IAM and GNHK datasets, highlighting the notable variety of simulated styles and capabilities of our method, are included in the supplementary material.

To assess and quantify the quality of the generated words, we compute the Fréchet Inception Distance (FID) score~\cite{dowson1982frechet}, the Mean Structural Similarity Index (MSSIM)~\cite{wang2004image}, the Root Mean Squared Error (RMSE), and the Learned Perceptual Image Patch Similarity (LPIPS)~\cite{zhang2018unreasonable} scores.
These metrics are commonly used to evaluate generative models. 
However, their use in HTG is not really intuitive, as they either rely on an ImageNet pre-trained network or compute pixel-wise similarities.
Furthermore, we approach the evaluation through a writer classification strategy, following a more document-oriented strategy.
To this end, we deploy a ResNet18 architecture~\cite{he2016deep}, pre-trained on ImageNet, and finetune it on the IAM database for the task of writer-style classification.
We use a \emph{different backbone} from our style extractor to avoid any bias induced by our model training.
The goal is to train the recognizer on a subset of the real training set and then evaluate its performance on the entire set of synthetic samples generated by different methods that simulate IAM (same words, same styles). 
This approach aims to determine whether the recognizer can correctly classify the generated samples, regardless of whether it has seen the corresponding real samples during training or not.

~\cref{tab:style_eval} presents the aforementioned metrics obtained using the different considered methods.
Our proposed method and its variations non-trivially outperform the other HTG approaches for all metrics.
Similarly, for the task of writer identification, the model successfully classifies a high percentage of samples generated by our method with an accuracy of 70.31\%.
This result overpasses the 68.25\% of the recent WordStylist approach that learns the style class in an explicit manner.
In general, WordStylist and DiffusionPen have very similar performances in most metrics.
This behavior is expected, as the two systems share backbone architecture. 
Our results suggest that, while we are able to reproduce the style slightly better than Wordstylist (which explicitly uses the style class as an embedding), we have managed to introduce more variation to the generated samples, which is the most crucial component in improving HTR performance when training on generated samples.
Within this experimental setup, we also conduct an ablation study on the usefulness of our proposed style extractor by exploring the role of the loss terms.
A breakdown of the loss terms of each ablation variation and additional discussion on the benefits of the style extractor are presented in the supplementary material.
Due to format issues, the metrics for VATr are not included in the table.
From the presented results, we can draw the following conclusions.
First, the writer classification accuracy is increased using the hybrid style embedding, namely through the joint classification and triplet scheme. 
This is not the case for the other metrics, but the differences are nonsignificant, and their relevance to the HTG task is far inferior compared to the writer classification paradigm.
Moreover, previous methods, such as GANwriting~\cite{kang2020ganwriting} and SmartPatch~\cite{mattick2021smartpatch}, despite having paved the way towards generating realistic images of handwritten words, seem unable to simulate the varieties of writing styles existing in IAM. 
Finally, our proposed hybrid style embedding outperforms all the reported methods for all the considered metrics.

\begin{table}[t]
  \centering
  \renewcommand{\arraystretch}{0.91}
    \addtolength{\tabcolsep}{3.5pt} 
    \caption{Comparison of FID, MSSIM, RMSE, LPIPS, and classification accuracy with previous methods. For FID, RMSE, and LPIPS, the lower, the better.}
    \scalebox{0.74}{
  \begin{tabular}{@{}lccccc@{}}
    \toprule
    \textbf{Method} & \textbf{FID\textdownarrow} & \textbf{MSSIM\textuparrow}  & \textbf{RMSE\textdownarrow}& 
    \textbf{LPIPS\textdownarrow}
    &\textbf{Acc(\%)\textuparrow}  \\
    \midrule
    Real IAM & --& -- &--&-- &$92.34$ \\
    \midrule
    GANwriting & $43.97$ &$0.777$&0.3118&0.2912&$\phantom{7}3.25$ \\
    SmartPatch &$50.21$ &$0.757$&0.3207&0.3003&$\phantom{7}2.14$ \\
    WordStylist & $34.20$&$0.955$ &0.1576&0.1080&$68.25$ \\
    \midrule
    DiffPen-class (Ours) & $22.23$&$0.967$&0.1587&0.1114&$68.04$ \\
    DiffPen-triplet (Ours) & $22.06$&$0.953$&0.1612&0.1127&$67.50$ \\
    DiffPen (Ours) & $22.54$ &$0.963$& 0.1505 & 0.1072 &$70.31$ \\
    \bottomrule
  \end{tabular}}
  \vspace{-.55cm}
  \label{tab:style_eval}
\end{table}

\vspace{-.5cm}
\noindent
\textbf{Unseen Styles.}
Unlike previous works that use Diffusion Models for HTG~\cite{zhu2023conditional, nikolaidou2023wordstylist}, DiffusionPen can imitate a writing style not seen during training from a few samples.
We present qualitative results of generating unseen writing styles in~\cref{fig:unseen_styles}.
Our method can replicate the style of the unseen style samples used as conditions and produce understandable text.
More generation examples of Unseen styles, with both IV and OOV words, are included in the supplementary material.
We should highlight the low number of required exemplars -- only $5$ -- used for the generation, enabling potential applications where writers' data are limited.
Additionally, the mean aggregation used over the exemplar embeddings suggests that one can use a variable number of exemplar images without issue.
We showcase how the number of samples in the few-shot setting affects the generation in the supplementary material.

\subsection{Handwriting Text Recognition}
\label{sub:htr}
Similar to previous works~\cite{nikolaidou2023wordstylist,zhu2023conditional}, we evaluate the quality of the generated handwritten text on the task of Handwriting Text Recognition (HTR) on the word level.
We use a CNN-LSTM HTR system~\cite{retsinas2022best} trained with Connectionist Temporal Classification (CTC) loss~\cite{graves2008novel}, as used in the evaluation process of~\cite{nikolaidou2023wordstylist}.

\begin{wraptable}{r}{5.5cm}
  \centering
  \renewcommand{\arraystretch}{0.95}
  \addtolength{\tabcolsep}{3.5pt}
  \vspace{-0.75 cm}
  \caption{Comparison of HTR Results using only the synthetic IAM samples for training. The closer to the Real IAM result (first row), the better.}
  \scalebox{0.72}{
  \begin{tabular}{@{}lcc@{}}
    \toprule
    \textbf{Dataset}  & \textbf{CER(\%)\textdownarrow} & \textbf{WER(\%)\textdownarrow}  \\
    \midrule
    Real IAM &$\phantom{3}5.16\pm0.01$&$14.49\pm0.07$ \\
    \midrule
    GANwriting  &$39.94\pm0.35$&$73.38\pm0.61$ \\
    SmartPatch  &$39.81\pm0.83$&$72.75\pm0.19$ \\
    VATr  &$21.74\pm0.32$&$50.55\pm0.47$ \\
    WordStylist  & $\phantom{3}8.26\pm0.05$&$23.36\pm0.16$ \\
    \midrule
    DiffPen-class (Ours) &$\phantom{3}7.12\pm0.03$&$18.55\pm0.10$ \\
    Diff-triplet (Ours)&$\phantom{3}7.13\pm0.11$&$18.48\pm0.14$ \\
    DiffPen (Ours)&\textbf{$\phantom{3}6.94\pm0.06$}&$18.11\pm0.25$ \\
    \bottomrule
  \end{tabular}}
  \vspace{-.5cm}
  \label{tab:htr_iam}
\end{wraptable}
\noindent

\textbf{Imitating IAM.}
We regenerate the training set and use the generated data to train the HTR system.
Then, we evaluate the HTR performance on the real test set, aiming to reach results as close as possible to the real training data.
The motivation behind this experiment is straightforward yet powerful. Specifically,
an HTG method could reproduce the performance of the real IAM data, or even surpass it, if these three abilities are satisfied: 1) the textual information is generated correctly, 2) styles differ substantially between them, and 3) given a text and a style a non-trivial variation would be generated.
Even if point (1) is very crucial, the main shortcomings of recent HTG methods concern points (2) and (3) in the sense that these methods do not generate enough variations in order to be efficiently utilized for such a learning task.
An example to understand the importance of this rationale is that if the inner-class variance of the generation process is trivial,
generating 10 times the common word ``and'', typically met numerous times in documents, can not provide any useful extra information to the training procedure. 

Concerning the imitation experiment, we follow the same steps in~\cite{kang2020ganwriting, mattick2021smartpatch, Bhunia_2021_ICCV, Pippi2023HandwrittenTG, nikolaidou2023wordstylist}. 
The HTR results are presented in Table~\ref{tab:htr_iam}.
Our method reaches the closest results to the real data with a CER of 6.94\% and WER of 18.11\%, outperforming all the other methods.
Similar to the quality assessment experiments, we include experiments to assess the effectiveness of our introduced style extractor and its loss terms.
The variations of DiffusionPen, where the style encoder is trained with the classification or the triplet protocol, achieve the next best performance. 
WordStylist achieves the next closest performance with a CER of 8.26\% and a WER of 23.36\%.

\noindent
\textbf{Improving HTR Performance.}
Given the results of~\cref{tab:htr_iam}, we use the data generated from the best performing method, which is our proposed DiffusionPen, as an augmentation to the real training set aiming to improve the performance of the baseline HTR system that achieves a CER of $5.16\%$ and a WER of $14.49\%$.
We also compare our performance with other works~\cite{Fogel2020ScrabbleGANSV, kang2020unsupervised, zhu2023conditional} that use synthetic data, as shown in in~\cref{tab:htr_iam_performance}.
Using the additional data from DiffusionPen can enhance the performance of the HTR system, showing promising potential for future use in larger generated datasets to assist the training process.

\begin{table}[t]
  \centering
  \renewcommand{\arraystretch}{0.91}
  \addtolength{\tabcolsep}{5.5pt}
     \caption{HTR performance with additional synthetic data to the real training set. The baseline values The baseline values are the ones from the original paper~\cite{retsinas2022best}.}
  \scalebox{0.77}{
  \begin{tabular}{@{}lccc@{}}
    \toprule
    \textbf{Dataset}  & \textbf{\# Synthetic Data} &\textbf{CER(\%)\textdownarrow} & \textbf{WER(\%)\textdownarrow}  \\
    \midrule
    ScrabbleGAN~\cite{Fogel2020ScrabbleGANSV} & 100K & 13.42& 23.61 \\
    Kang et al.~\cite{kang2020unsupervised}& - & 6.75& 17.76 \\
    CTIG-DM~\cite{zhu2023conditional} & 1M &5.19 & 13.37 \\
    \midrule
    Baseline~\cite{retsinas2022best} & - &$5.14$ & $14.33$ \\
    DiffusionPen (Ours) & 55K&4.71&13.61 \\
    \bottomrule
  \end{tabular}}
  \label{tab:htr_iam_performance}
\end{table}

\vspace{-1.5mm}
\subsection{Style Variation}

To reflect the natural variations of human handwriting in automatic handwriting generation, it is crucial not only to generate realistic text but also to have diversity. 
Under our framework, style variation can be simulated seamlessly. 
First, the few-shot paradigm is, by its nature, a variation-promoting mechanism since, for the same style, different embeddings are calculated from the randomly selected exemplar images.
Nonetheless, the style space is less sensitive to ``exploration'', compared to previous works, enabling us to search for more ``aggressive'' style augmentations. 
Specifically, we explore the aspect of style variation through the following scenarios: interpolation and noisy style embedding.

\begin{figure}
\begin{center}
\centering
\begin{subfigure}[t]{0.91\textwidth}
    \centering
    \renewcommand{\arraystretch}{0.9}
    \addtolength{\tabcolsep}{1.5pt} 
    \scalebox{0.82}{
    \begin{tabular}{ccccccc}
      \includegraphics[scale=0.16]{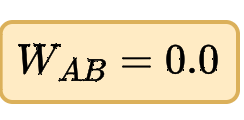} &  \includegraphics[scale=0.16]{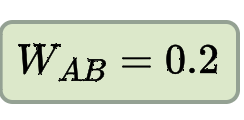} & \includegraphics[scale=0.16]{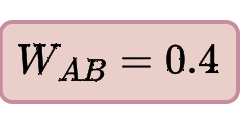} & \includegraphics[scale=0.16]{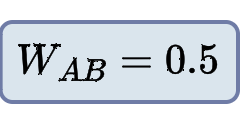} &\includegraphics[scale=0.16]{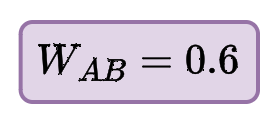} & \includegraphics[scale=0.16]{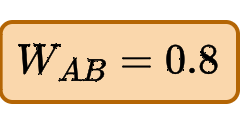}& \includegraphics[scale=0.16]{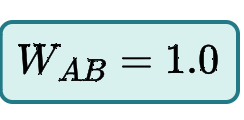} \\
     \includegraphics[scale=0.15]{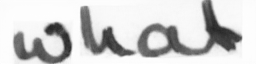} & 
     \includegraphics[scale=0.15]{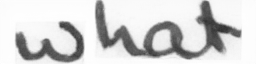}  & 
     \includegraphics[scale=0.15]{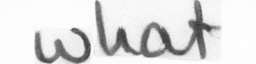}& 
     \includegraphics[scale=0.16]{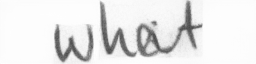} & 
     \includegraphics[scale=0.16]{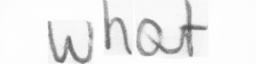}  & 
     \includegraphics[scale=0.16]{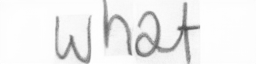} & 
     \includegraphics[scale=0.16]{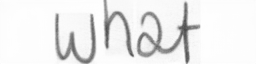} 
     \\
     \includegraphics[scale=0.16]{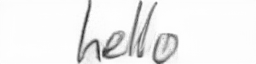} & 
     \includegraphics[scale=0.16]{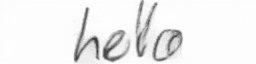}  & 
     \includegraphics[scale=0.16]{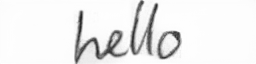} &  
     \includegraphics[scale=0.16]{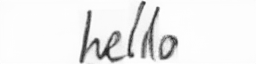}&
     \includegraphics[scale=0.16]{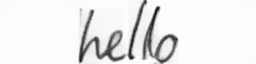}  & 
     \includegraphics[scale=0.16]{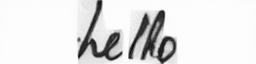} & 
     \includegraphics[scale=0.16]{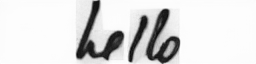} 
    \\
     \includegraphics[scale=0.16]{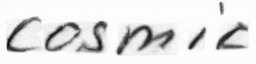} & 
     \includegraphics[scale=0.16]{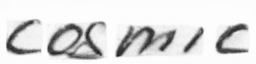}  & 
     \includegraphics[scale=0.16]{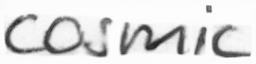} &  
     \includegraphics[scale=0.16]{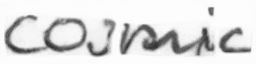}&
     \includegraphics[scale=0.16]{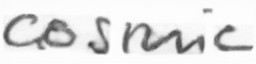}  & 
     \includegraphics[scale=0.16]{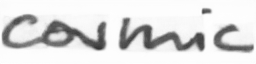} & 
     \includegraphics[scale=0.16]{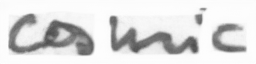} 
     \\
     \includegraphics[scale=0.16]{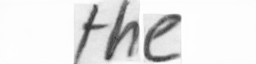} & 
     \includegraphics[scale=0.16]{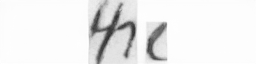}  & 
     \includegraphics[scale=0.16]{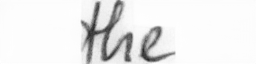} &  
     \includegraphics[scale=0.16]{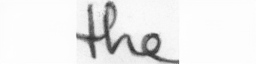}&
     \includegraphics[scale=0.16]{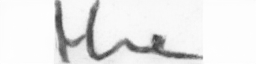}  & 
     \includegraphics[scale=0.16]{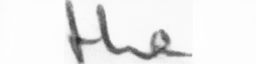} & 
     \includegraphics[scale=0.16]{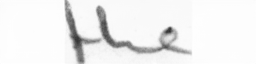} 
     
    \end{tabular}}
\caption{Interpolation between two writer styles with various values of the weight $W_{AB}$.}
\label{fig:interpolation}
\end{subfigure}

\begin{subfigure}[t]{\textwidth}
    \centering
    \vspace{0.2cm}    \includegraphics[width=0.73\linewidth]{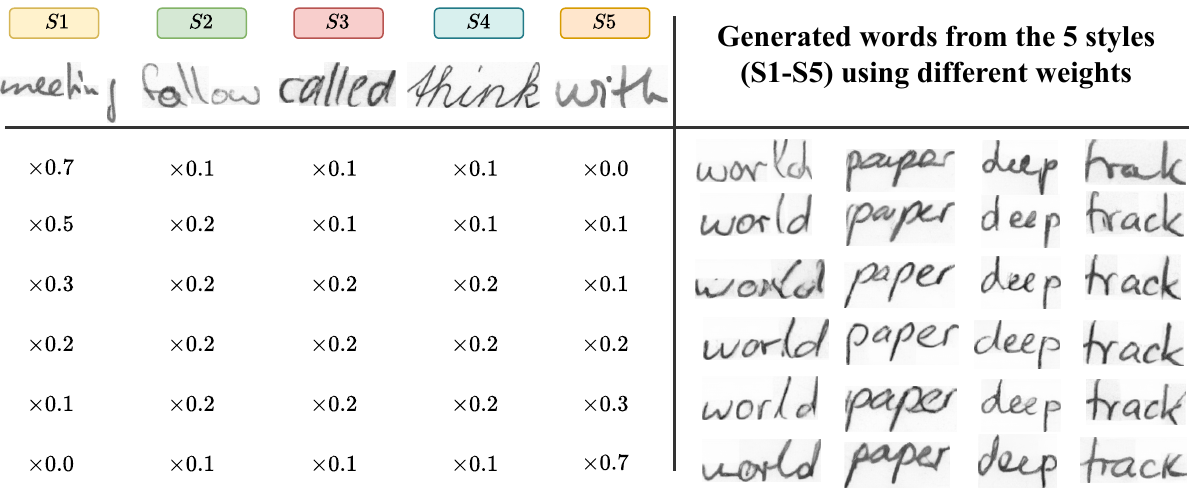}
    \caption{Multi-style mixture generation.}
    \label{fig:multistyle}
\end{subfigure}
\caption{Generated samples by \textbf{(a)} interpolation and \textbf{(b)} multiple styles.}
\vspace{-.7cm}
\label{fig:interpol}
\end{center}
\end{figure}

\noindent
\textbf{Interpolation.}
We tweak the style embedding between two writer styles to obtain samples between the two styles.
We interpolate a generated sample of a style $S_A$ to a style $S_B$ using a weighted average \(S_{AB}=(1-W_{AB})S_A + W_{AB}S_B\) for different values of $W_{AB}$.
~\cref{fig:interpolation} presents several examples of interpolating between two random styles.
We can see from the results that there is a smooth transition between the styles as the value of $W_{AB}$ changes.
An interesting result can be observed in the last row of the figure, where for $W_{AB}=0.5$ of the word \texttt{the}, the curvature of the character \texttt{h} does not resemble either of the style classes. 
This hints that we may ``stumbled'' upon an entirely different style by cursing through the style space. 

\noindent
\textbf{Multi-style Mixture.}
Following the interpolation strategy, instead of mixing two writing styles, we generate samples by conditioning on 5 different styles.
~\cref{fig:multistyle} shows that combining 5 different styles (S1-S5) with different weights can create new styles and variations of the same word, showcasing our model's capability of exploiting the style space.
More style mixture examples are presented in the supplementary material.

\noindent
\textbf{Noisy Style Embedding.}
We explore two different noise variations that could inject diversity into the generated data.
One variation is the \textit{noisy style embedding}, where we add random noise to the style embedding to avoid explicitly getting the learned style.
The other variation is the \textit{noise bias}.
We inject noise bias into the diffusion model by replacing the random noise given to the diffusion model as the initial latent variable with a noisy image from a wished style.
We regenerate the IAM database using either the noise bias to initiate the denoising of the model or the noisy style embedding for different levels of noise or the combination.
Then, we train the HTR system using the different generated databases instead of the real IAM training set to determine how the noise injections can assist the data variation in the generation.
\begin{wraptable}{r}{6.3cm}
  \centering
  \renewcommand{\arraystretch}{0.91}
  \addtolength{\tabcolsep}{4.5pt}
  \vspace{-.2cm}
  \caption{Exploration of random noise variations in the style embedding or the prior. The \checkmark indicates whether there is a bias in the prior, and the values 0.25-2.00 indicate the weight of the noise added to the style embedding.}
  \scalebox{0.71}{
  \begin{tabular}{@{}cccc@{}}
    \toprule
    \textbf{Noise Bias}  & \textbf{Style Noise} & \textbf{CER(\%)\textdownarrow} & \textbf{WER(\%)\textdownarrow}  \\
    \midrule
    -- & 2.00&$7.56\pm0.06$&$19.56\pm0.11$ \\
    -- & 0.50 &$6.99\pm0.06$&$18.30\pm0.21$ \\
    -- & 0.25&$6.79\pm0.08$&$17.85\pm0.09$\\
    \checkmark & --&$6.93\pm0.20$&$18.18\pm0.49$\\
    \checkmark & 0.25 &$7.02\pm0.20$&$18.26\pm0.26$\\
    \bottomrule
  \end{tabular}}
  \vspace{-2.6mm}
  
  \label{tab:htr_variation}
\end{wraptable}

\noindent
The results are presented in~\cref{tab:htr_variation}.
One can observe that a small magnitude of noise (i.e., $0.25$) could assist the training procedure and improve the HTR performance, indicating that the noise variations are beneficial. 
On the other hand, increasing noise above a threshold may complicate the system - potentially diverging from the manifold of useful styles. 
Finally, the noise bias does not seem to assist performance, even when combined with the noisy embedding of $0.25$ magnitude.

\section{Limitations and Ethical Considerations}
\label{sec:limitations}
\begin{figure}
    \vspace{-.1cm}
    \begin{subfigure}[t]{0.42\textwidth}
        \centering
        \includegraphics[width=0.7\linewidth]{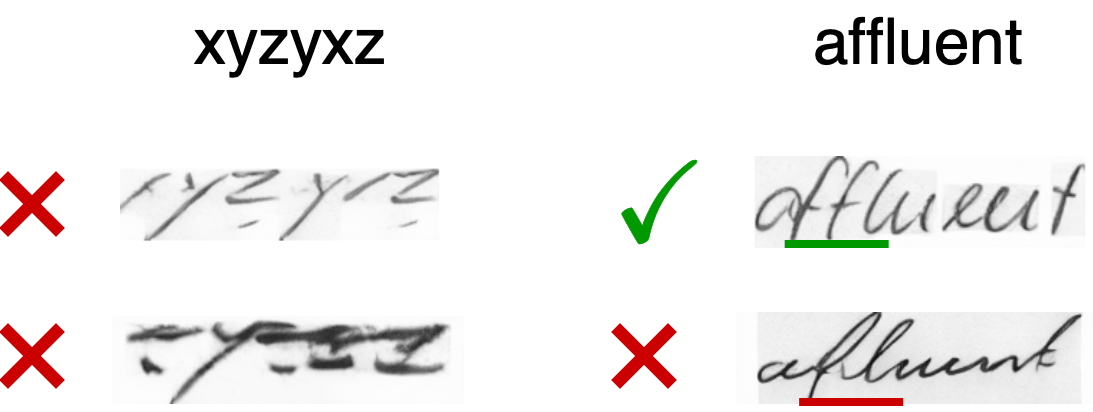}
        \caption{Rare character combinations (left) and complex ligatures (right).}
        \label{fig:xyz}
    \end{subfigure}
    \hfill
    \begin{subfigure}[t]{0.42\textwidth}
        \centering
        \includegraphics[width=0.9\linewidth]{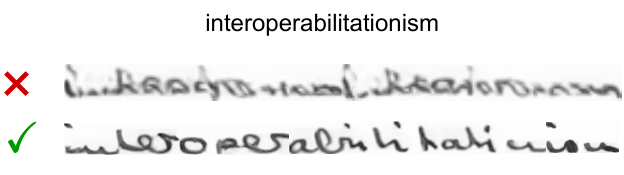}
        \caption{Unsuccessful generation of the word \texttt{interoperabilitationism} and solution.}
        \label{fig:interoperabilitationism}
    \end{subfigure}
    \caption{Examples of fail cases. See text for details.}
    \label{fig:limitations}
\end{figure}

\textbf{Limitations.}
Fail cases may occur in rare combinations of characters (``xyzyxz") or complex ligatures in some cursive styles (``affluent"), as shown in~\cref{fig:xyz}.
Considering the word ``affluent", our model successfully generates the top style that has less complex connections, while it struggles with the cursive ``ffl" ligature in the more complex style on the bottom.
This might not be observed in comparing GAN-based methods, which tend to simplify the style to force the generation of understandable text.
However, a trade-off between text and style variation should be found to get a robust generation.
Furthermore, although our method is able to generate words over 10 characters (see~\cref{fig:10_chars}), there is still a length limit due to the maximum word length present in the training and the noise initialization of the denoising process.
Such a case is presented in~\cref{fig:interoperabilitationism}, where the model fails to generate the word ``interoperabilitationism" (top).
This issue could be solved by practical tricks such as patching generated samples of smaller parts of the word, as presented in the bottom of~\cref{fig:interoperabilitationism}, where we generate the parts ``interoper", ``abilitation" and ``ism". 

\noindent
\textbf{Ethical Considerations.}
The possibility of mimicking a specific handwriting style from a limited set of image samples poses a significant risk for handwriting forgery.
Although technologically impressive, this capability of HTG models could potentially be exploited for fraud and identity theft by creating unwanted text or signatures resembling a person's writing style.
There is an entire field in forensics that attempts to detect such frauds with predictive models and techniques to distinguish authentic from machine-generated imitations.
\section{Conclusion}
\label{sec:conclusion}

We presented \emph{DiffusionPen}, a few-shot style latent diffusion model for handwriting generation that incorporates a hybrid metric-learning and classification-based style extractor. 
Our approach captures stylistic features
of seen and unseen writers while preserving readable text
content. 
We present qualitative and quantitative results and compare them with other SotA methods based on GANs, Transformers, and DDPM.
Given the HTR task results, our method is the closest to the performance of the real IAM, and using data generated from \emph{DiffusionPen} enhances HTR performance, allowing us to envisage utilizing HTG large-scale dataset generation.
Finally, further exploration of style and noise variation in different stylistic aspects shows potential directions for future work. 

\newpage 

\section*{Acknowledgment}
\noindent
The computations and data handling were enabled by the Berzelius resource provided by the Knut and Alice Wallenberg Foundation at the National Supercomputer Centre at Linköping University.
The publication/registration fees were partially covered by the University of West Attica.

%

\bibliographystyle{splncs04}
\bibliography{main}


\clearpage  
\pagestyle{plain}
        \begin{center}
    \LARGE\bfseries DiffusionPen: Towards Controlling the Style of Handwritten Text Generation Supplementary Material
    \vspace{1.5em}
\end{center}

\appendix
\counterwithin{figure}{section}
\counterwithin{table}{section}
\renewcommand\thefigure{\thesection\arabic{figure}}
\renewcommand\thetable{\thesection\arabic{table}}

We present the supplementary material for our proposed few-shot styled Handwritten Text Generation system, named DiffusionPen (DiffPen).
In particular, we include additional information considering the Style Encoder backbone and the versions of our method in~\cref{sec:style_sup}.  
Furthermore, we present additional qualitative results of DiffusionPen in~\cref{sec:qualitative_sup} and compute the FID score for the generated IAM test sets that combine both In-Vocabulary (IV) and Out-of-Vocabulary (OOV) words but only include Unseen styles.
In the same section, we showcase the ability of the model to generate smaller paragraphs or longer words.
In~\cref{sec:style_variation_sup}, we present further visual examples to explore the effect of the style embedding, the noise induction in the style embedding, and the initialization noise bias. 
Furthermore, we show examples of style mixtures where we generate new styles by combining up to 5 different styles.
Moreover, we show that our model can generate high-quality samples even with 1-shot style sampling.
Finally, \cref{sec:htr_sup} shows extended Handwriting Text Recognition (HTR) results using generated data as an augmentation to the real IAM data~\cite{791885} to improve the HTR performance.

\section{Style Encoder}
\label{sec:style_sup}

\textbf{Backbone.}
In our work, we utilize a Style Encoder $S_E$ that is trained with a hybrid triplet and classification loss to model the style of the word images. 
We conducted preliminary experiments on style classification using ResNet18, ResNet50, and MobileNetV2, all pre-trained on ImageNet, to choose the backbone of the Style Encoder.
The resulting accuracy, along with the number of parameters of every model, is presented in~\cref{tab:style_ablation_sup}.
One can see that the performance of all three models is similar, with ResNet50 achieving the best accuracy, being close to the following best, which is MobileNetV2.
However, both ResNet architectures are much heavier in terms of parameters than the MobileNet architecture.
Thus, MobileNet's combination of high performance and lightweight design made us proceed with that choice.

\begin{table}[ht]
  \centering
  \setlength{\tabcolsep}{15pt}
  \caption{Ablation on the Style Encoder backbone network. }
  \begin{tabular}{@{}lcc@{}}
    \toprule
    \textbf{Method} & \textbf{Accuracy(\%)\textuparrow} &\#parameters\\
    \midrule

     MobileNetV2 &  89.21 &2.7M\\
    ResNet18 & 88.23 & 11.4M\\
   
    
    ResNet50  & 90.37 & 24.2M\\
    \bottomrule

  \end{tabular}
  
  \label{tab:style_ablation_sup}
\end{table}

\noindent
\textbf{Hybrid Loss.}
Within the experimental setup of our work, we conduct an ablation on the usefulness of our proposed style extractor by exploring the role of the loss terms.
A breakdown of the loss terms of every ablation variation is presented in~\cref{tab:style_ablation_loss_sup}.
DiffPen corresponds to the model that uses the hybrid Style Encoder with both triplet and classification terms in the loss.
Besides the results presented in the full model, we also present qualitative results using the model that uses each loss term separately.
Hence, DiffPen-class corresponds to the model where the Style Encoder is trained only with Cross-Entropy loss $\mathcal{L}_{class}$ and DiffPen-triplet $\mathcal{L}_{triplet}$ to the one trained only with the triplet loss.

In general, a simple classification loss (as used in previous approaches) will indeed generate samples that look like specific styles. 
However, such an approach forms a space of style descriptors that can easily degenerate to a set of centers around which points are classified to a style in a ``nearest neighbour'' sense - this explains the limited style variability of previous methods. 
A classification loss is adequate for a \emph{discriminative} model, but here it is inadequate because it is oblivious of the topology of the inferred space.
The proposed loss faces this problem exactly by explicitly bringing into play \emph{the metric characteristics} of the inferred style space.
In practical terms, while the result in Tab1 doesn't seem significant (from WordStylist to DiffPen), there is still an improvement of $~2\%$ in reproducing the styles.
The improvement is also very clear in Table 2 of the main paper, where if we run a significance test between DiffPen-class and DiffPen, we obtain a t-value of 4.77 and a p-value of 0.0088, making the improvement significant. 
All our model variations are also significantly better than the baseline WordStylist.
This proves that we significantly add more variation to our produced samples, which is the main goal of our paper.

\begin{table}[t]
  \centering
  \caption{Loss terms included in the variations of our proposed method for the ablation of the style encoder $S_E$. $\mathcal{L}_{class}$ represents the classification loss term and $\mathcal{L}_{triplet}$ the metric-learning term.}
  \begin{tabular}{@{}lcc@{}}
    \toprule
    \textbf{Method}  & $\mathcal{L}_{class}$  & $\mathcal{L}_{triplet}$\\
    \midrule
    DiffPen-class & \checkmark&\\
    DiffPen-triplet &&\checkmark\\
    DiffPen &\checkmark&\checkmark\\
\bottomrule
  \end{tabular}

  \label{tab:style_ablation_loss_sup}
\end{table}

\section{Qualitative and Quantitative Results}
\label{sec:qualitative_sup}

\begin{figure*}
\centering
\includegraphics[width=0.95\textwidth]{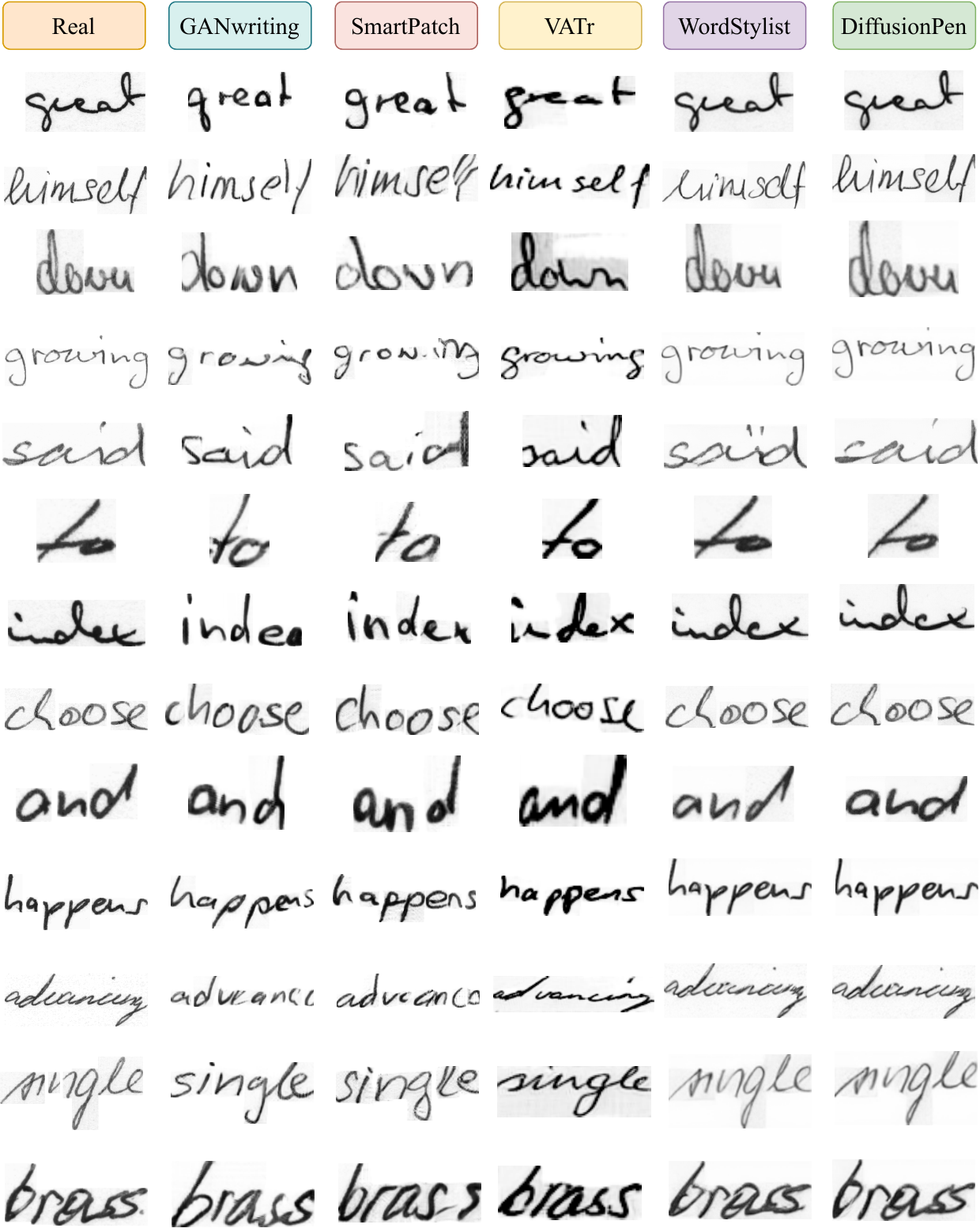}
    \caption{Qualitative results of In-Vocabulary (IV) words and Seen (S) styles. We compare our method with GANwriting, SmartPatch, VATr, and WordStylist.}
    \label{fig:qualitative_sup}
\end{figure*}

\noindent
\textbf{Comparison wit SotA.}
We present additional qualitative generated examples using our method and the comparing methods GANwriting~\cite{kang2020ganwriting}, SmartPatch~\cite{mattick2021smartpatch}, Visual Archetype Transformers (VATr)~\cite{Pippi2023HandwrittenTG}, and WordStylist~\cite{nikolaidou2023wordstylist}.
~\cref{fig:qualitative_sup} shows several In-Vocabulary words of seen writer styles present in the original train set, created using the different generative methods.
To quantify the observed results, we compute the FID score between the real and the generated test set, which contains solely unseen styles and both IV and OOV words, and compare with the corresponding test sets generated using GANwriting and SmartPatch.
The resulting FID scores are presented in~\cref{tab:fid_sup}, where we can see that our system achieves the best result on the generated test set.
It should be noted that we cannot compute the test set FID for WordStylist, as the system can only generate seen styles.
The FID scores obtained from VATr are notably higher than the rest of the methods, and the cause of that requires further investigation.
Thus, similar to our main paper, we decide not to include the FID for the VATr method.
\begin{table}[t]
  \centering
  \setlength{\tabcolsep}{15pt}
  \caption{Comparison of FID with previous GAN-based methods on the generated test set of IAM database. The test set consists of only Unseen styles and IV and OOV words. For FID, the lower, the better. }
  \begin{tabular}{@{}lc@{}}
    \toprule
    \textbf{Method} & \textbf{FID\textdownarrow} \\
    \midrule
    SmartPatch &$48.01$  \\
    GANwriting & $44.67$ \\
    DiffusionPen (Ours)  & $29.77$ \\
    \bottomrule

  \end{tabular}
  
  \label{tab:fid_sup}
\end{table}

\begin{figure*}
\centering
\includegraphics[width=1\textwidth]{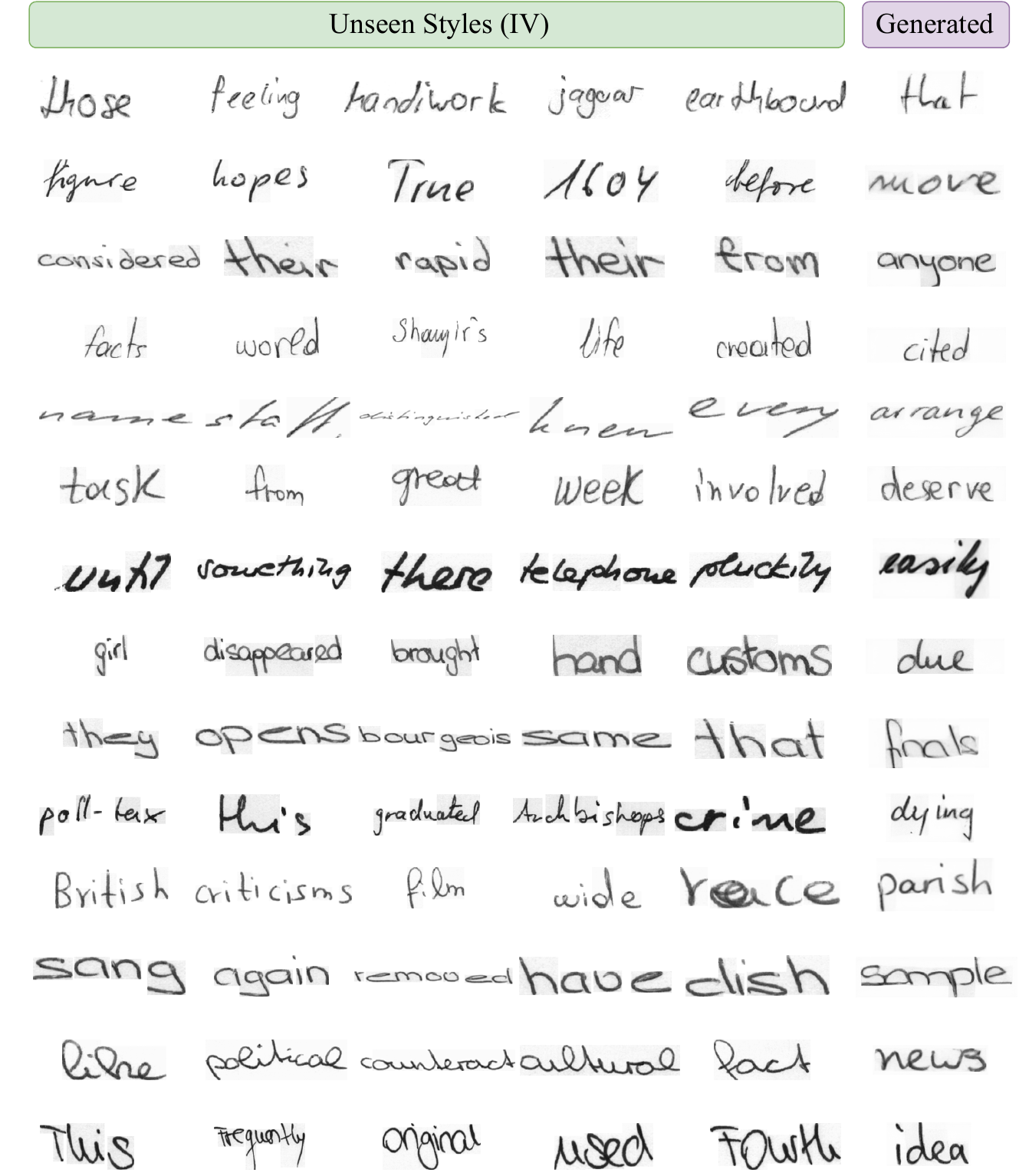}
    \caption{Qualitative results of In-Vocabulary (IV) words and Uneen (U) styles. The left column (Unseen Styles IV) shows the style samples used for the 5-shot condition, and the right column (Generated) shows the generated IV word.}
    \label{fig:unseen_styles_sup}
\end{figure*}

\begin{figure*}
\centering
\includegraphics[width=0.98\textwidth]{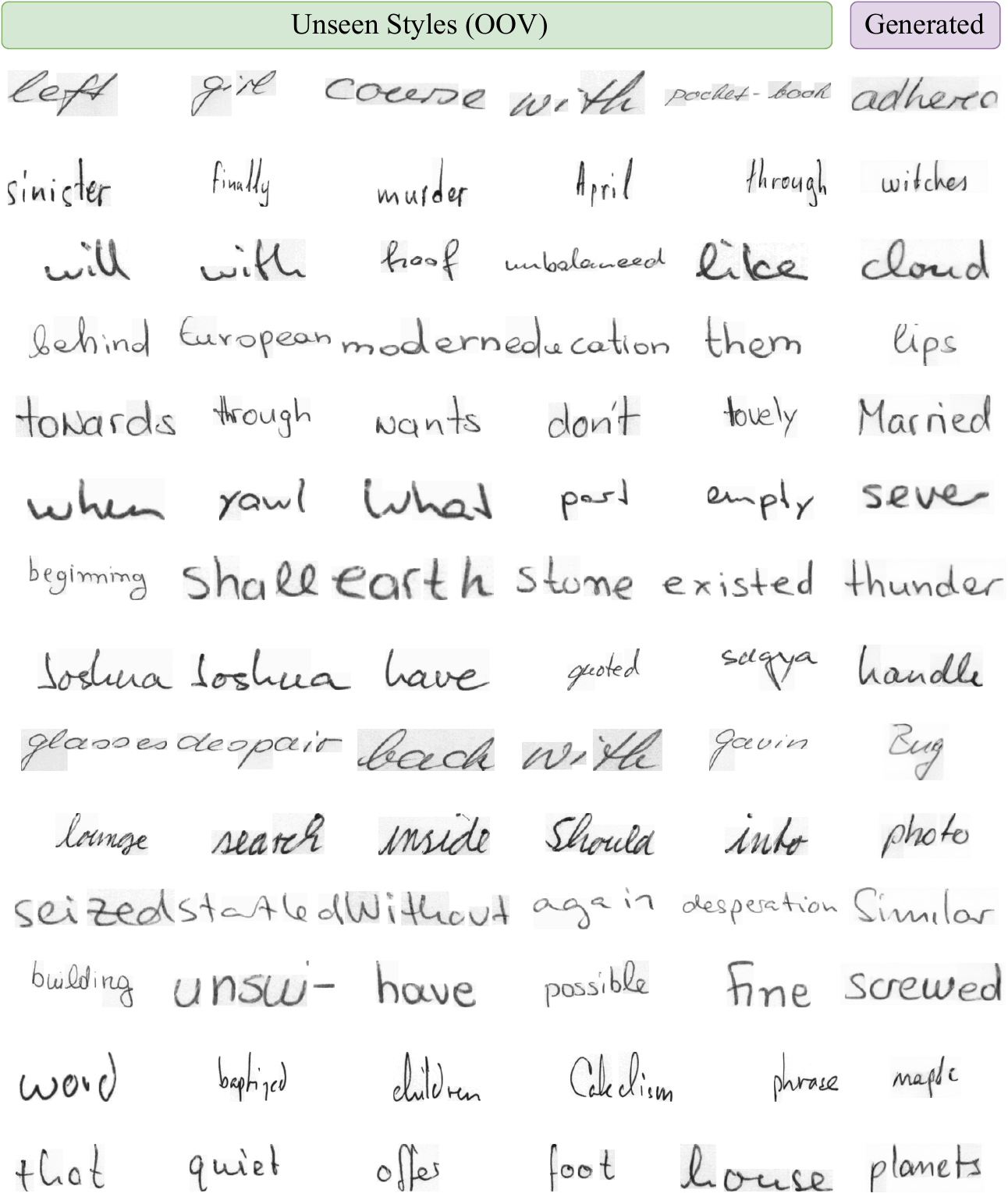}
    \caption{Qualitative results of Out-of-Vocabulary (OOV) words and Uneen (U) styles. The left column (Unseen Styles OOV) shows the style samples used for the 5-shot condition, and the right column (Generated) the generated OOV word.}
    \label{fig:oov_unseen_styles_sup}
\end{figure*}

\noindent
\textbf{Unseen Styles.}
Furthermore, we present additional qualitative results of unseen styles that were not present during the training of our proposed system for both In-Vocabulary (IV) and Out-of-Vocabulary (OOV) words.
~\cref{fig:unseen_styles_sup} shows the generated IV words on the right column (Generated) and the unseen style samples on the right column that constitute the 5-shot style embedding (Unseen Styles - IV).
~\cref{fig:oov_unseen_styles_sup} shows similar results for the case of Unseen styles and OOV words (Unseen Styles - OOV).
In both cases, one can observe that the generated handwriting on the right column maintains a consistent style in terms of letter formation, spacing, slant, and thickness with the style samples, suggesting that our method has effectively learned the handwriting characteristics from the limited set of unseen examples provided as a condition. 

\begin{figure*}[t]
\centering
\includegraphics[width=0.99\textwidth]{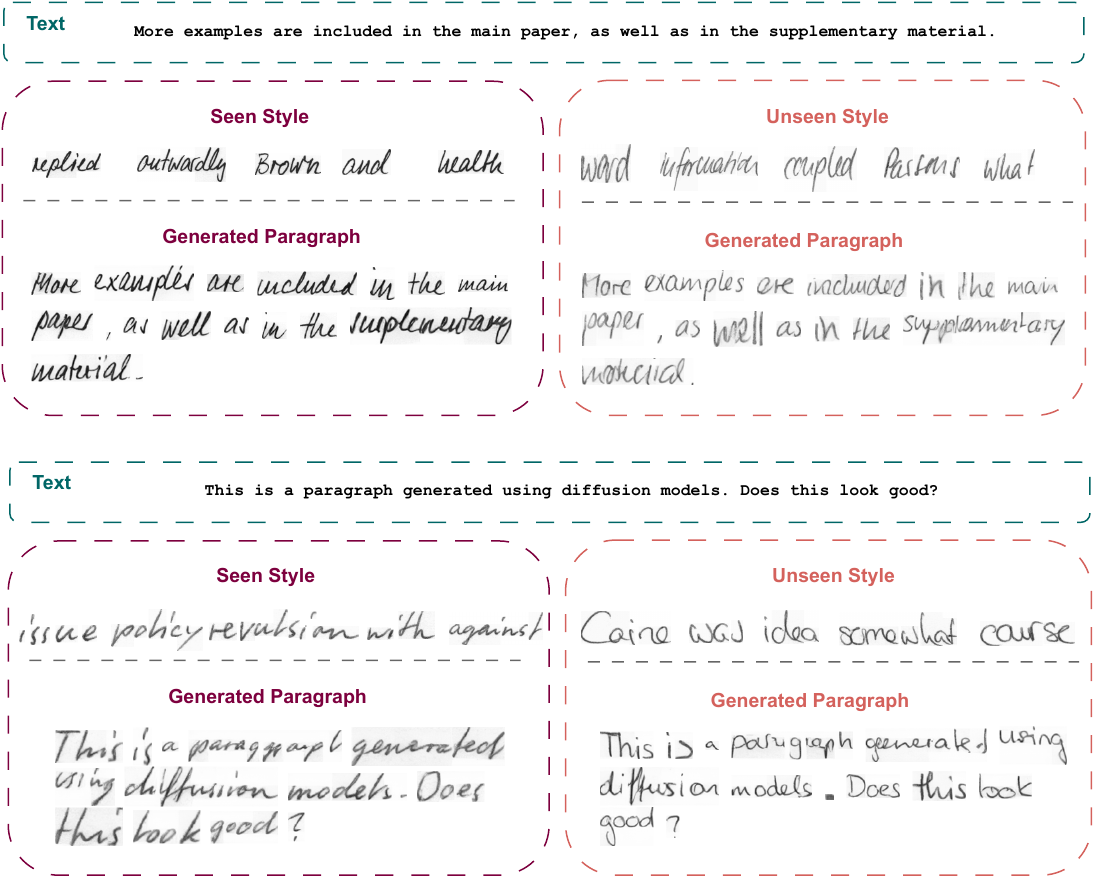}
    \caption{Small paragraphs generated in a Seen (left) and Unseen Style (right).}
    \label{fig:paragraphs_supplementary}
\end{figure*}

\noindent
\textbf{Paragraphs.}
We present two small paragraphs, comprised of two sentences, generated using our method in~\cref{fig:paragraphs_supplementary}.
This way, we show the practical applicability of not constraining the generation in words and the ability to generate larger parts of text by using 5 style word samples and specific content as conditions.

\begin{figure*}[t]
\centering
\includegraphics[width=0.9\textwidth]{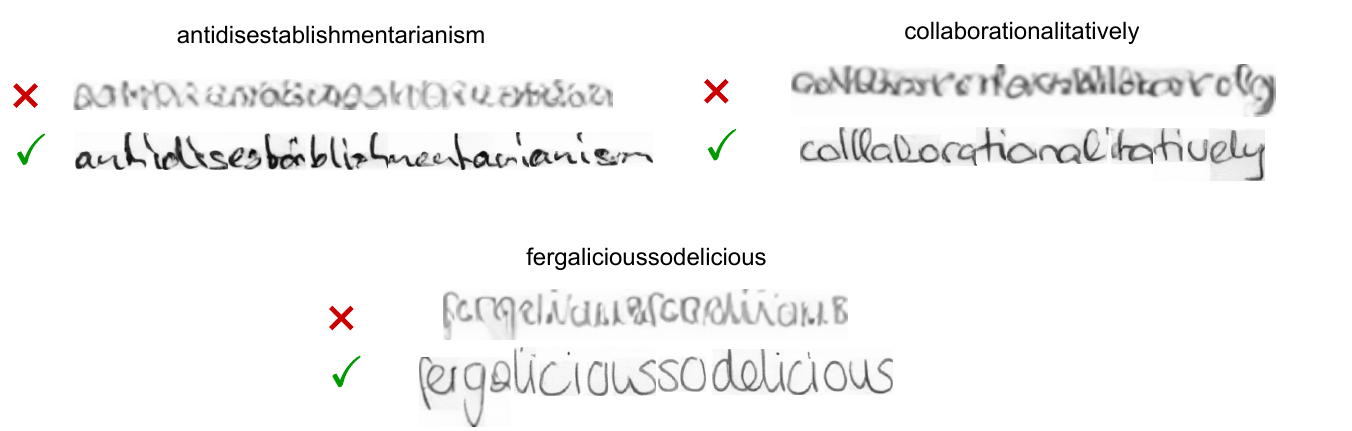}
    \caption{Generation of very long words.}
    \label{fig:limit_supplementary}
\end{figure*}

\noindent
\textbf{Limitations.}
As showcased in the main paper, our method has the constraint of generating the words in a specific image size due to noise initialization during sampling.
Furthermore, the dataset is limited to a maximum number of characters in the training set, which limits the model when asking the model to generate longer words.
However, this can be solved by patching smaller parts of the long word.
We present a few more examples in~\cref{fig:limit_supplementary}.
In these examples, we manage to generate the word ``antidisestablishmentarianism" through the generation of the words ``antidis", ``establish", ``mentarianism", and ``anism".
Similarly, we generate the word ``collaborationalitatively" through the combination of ``collabora", ``tionalita", and ``tively" and the word ``fergalicioussodelicious" through the generation of the subwords ``ferga", ``licious", ``so", and ``delicious".
In the presented examples, we have used a random split of the long words, as the main point is to have subparts shorter than the max word length present in the dataset. 
A systematic strategy could be devised by breaking a long word into (randomly-lengthed) segments, with each segment length smaller than the maximum word length.

\begin{figure}[ht]
  \centering
\includegraphics[width=0.7\linewidth]{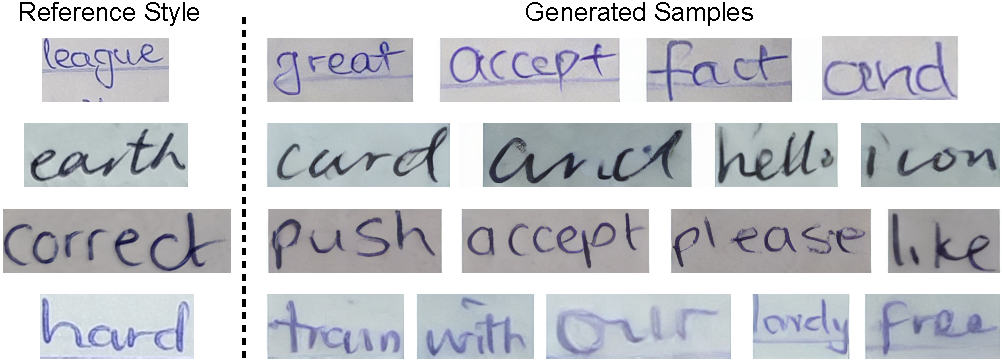}
   \caption{GNHK generated examples using DiffusionPen.}
   \label{fig:gnhk_sup}
\end{figure}

\noindent
\textbf{GNHK dataset.}
We include several qualitative results of the GNHK dataset~\cite{lee2021gnhk} on the word level in~\cref{fig:gnhk_sup}.
The figure shows a reference style and the corresponding generated samples of the randomly selected conditioned text.
We can see that our method successfully generates samples imitating the GNHK style.
However, the dataset is much more complex than the IAM database, with more complex backgrounds and less benchmarking on the word level as it is provided as page images.
Hence, more experimentation and adaptation are needed to meet the dataset's needs.

\section{Style Variation}
\label{sec:style_variation_sup}
We explore the effect of the style embedding on the generation process.
We visualize a few examples of the same word-writer pair generated multiple times using different style embedding conditions.
Furthermore, we explore the noise induction to the style embedding as well as the effect of adding bias to the prior noise that initializes the sampling process of the diffusion model. 

\begin{figure*}
\centering
\includegraphics[width=0.92\textwidth]{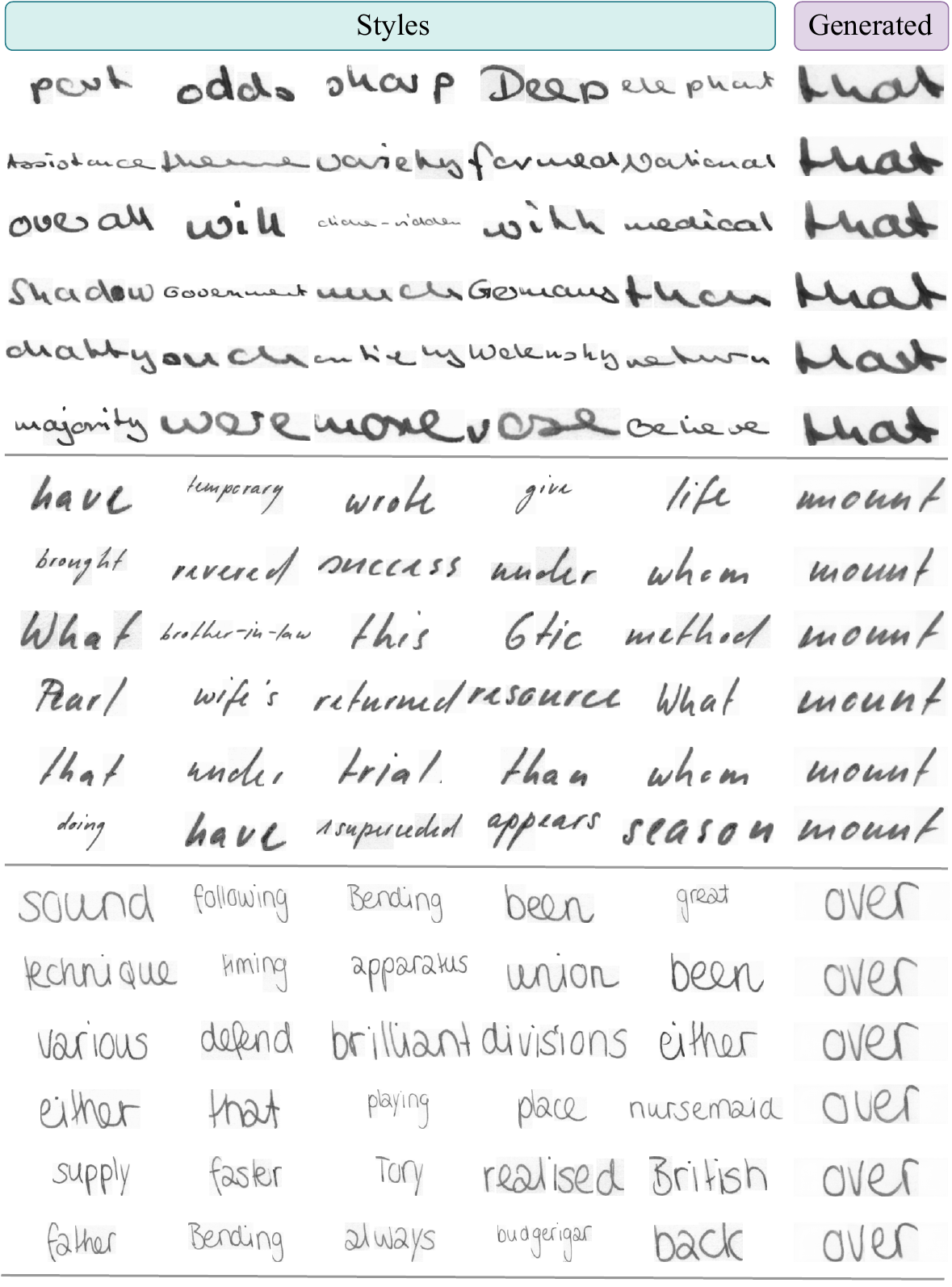}
    \caption{Multiple generations of the same word (right column), conditioning on different seen style samples (left column). For every different style combination, we get a different variation of the word.}
    \label{fig:style_effect_sup}
\end{figure*}

\noindent
\textbf{Style Embedding.}
~\cref{fig:style_effect_sup} shows the exploration of the style embedding.
For a fixed text content, we generate the same word written by the same writer style multiple times while changing the style samples that constitute the style embedding condition.
The results show that, for different style examples as conditions, the generated word has different variations.

\begin{figure*}
\centering
\includegraphics[width=\textwidth]{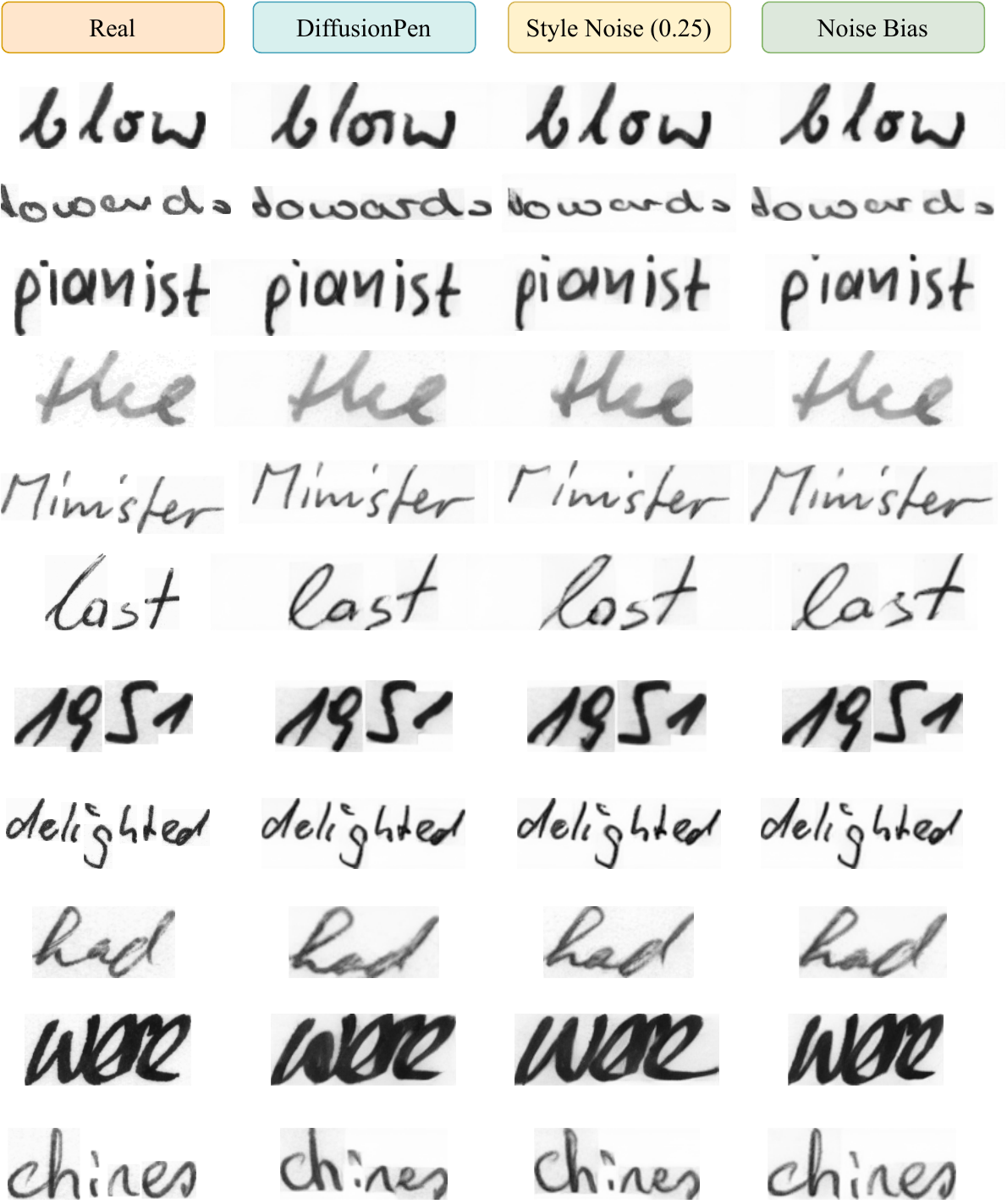}
    \caption{Style variations of the noisy style embedding and the noise bias exploration.}
    \label{fig:style_variation_sup}
\end{figure*}

\noindent
\textbf{Noisy Style Embedding \& Noise Bias.}
Following our experiments presented in the main paper, we continue the exploration of the style variation through noise induction to the style embedding or through biasing the initialization noise.
In~\cref{fig:style_variation_sup}, we present qualitative results and compare the generation of the same samples with our method (DiffusionPen) without any change, with the noise induction to the style condition embedding with a magnitude of 0.25 (Style Noise 0.25), and with the prior noise bias, where instead of initializing the generation with random noise, we randomly select an image from the same style and adding noise to it (Noise Bias). 
While the results are very close between the different cases, having a closer look, one can observe differences.
The noise induction (Style Noise 0.25) seems to generate some marks in character ``t" of the word ``towards" or the digit ``9" in ``1951" that resemble ink stains.
Furthermore, while all cases have the accent of the first ``i" of the word ``pianist" slightly on the left of the character, the Noise 0.25 case is placed right on top of the letter.
Considering the Noise Bias case (last column), the generated results seem to be the closest to the real data, such as the words ``were", ``1951", and ``chines". There are small details that differentiate them, such as the extended ``r" in ``Minister", or the ``b" in ``blow".
These details can induce small variations in the data while striving to control the generation's style.
We further explore the HTR performance of these cases and comment on the results in~\cref{sec:htr_sup}.

\begin{table}[ht]
  \centering
  \setlength{\tabcolsep}{3pt}
  \caption{HTR performance with additional synthetic data to the real training set. The first row shows the performance of the real IAM database without any augmentation. The second row shows the performance of the real IAM dataset with the additional IAM synthetic samples generated from DiffusionPen. The results show the mean and standard deviation over three runs of each experiment.}
  \begin{tabular}{@{}lcc@{}}
    \toprule
    \textbf{Dataset}  & \textbf{CER (\%) \textdownarrow} & \textbf{WER (\%) \textdownarrow}  \\
    \midrule
    Real IAM &$5.16\pm0.01$&$14.49\pm0.07$ \\
    \midrule
    Real IAM + GANwriting IAM&$5.22\pm0.03$&$14.40\pm0.13$ \\
    Real IAM + SmartPatch IAM&$5.48\pm0.13$&$14.97\pm0.35$ \\
    Real IAM + VATr IAM&$5.20\pm0.16$&$14.37\pm0.40$ \\
    Real IAM + WordStylist IAM&$4.75\pm0.04$&$13.29\pm0.11$ \\
    Real IAM + DiffPen IAM &$4.78\pm0.07$&$13.72\pm0.13$ \\

    \bottomrule
  \end{tabular}

  \label{tab:htr_sup}
\end{table}

\begin{table}[ht]
  \centering
  \setlength{\tabcolsep}{3pt}
  \caption{HTR performance of the real IAM data and the data generated from DiffusionPen using the two exploration variations. Style Noise (0.25) represents the synthetic data created by adding noise to the style embedding of 0.25 magnitude, while Noise Bias represents the data where instead of random noise, a noisy image belonging to the same writer as the style condition is used to initialize the sampling process.}
  \begin{tabular}{@{}lcc@{}}
    \toprule
    \textbf{Dataset}  & \textbf{CER (\%) \textdownarrow} & \textbf{WER (\%) \textdownarrow}  \\
    \midrule
    Real IAM &$5.16\pm0.01$&$14.49\pm0.07$ \\
    \midrule
    Real IAM + Style Noise (0.25) &$4.88\pm0.05$&$13.97\pm0.20$ \\
    Real IAM + Noise Bias&$4.86\pm0.02$&$13.72\pm0.16$ \\

    \bottomrule
  \end{tabular}
  \label{tab:htr_style_sup}
\end{table}

\begin{figure*}[ht]
  \centering
  
  \begin{subfigure}{0.9\textwidth}
        \centering
        \includegraphics[width=\linewidth]{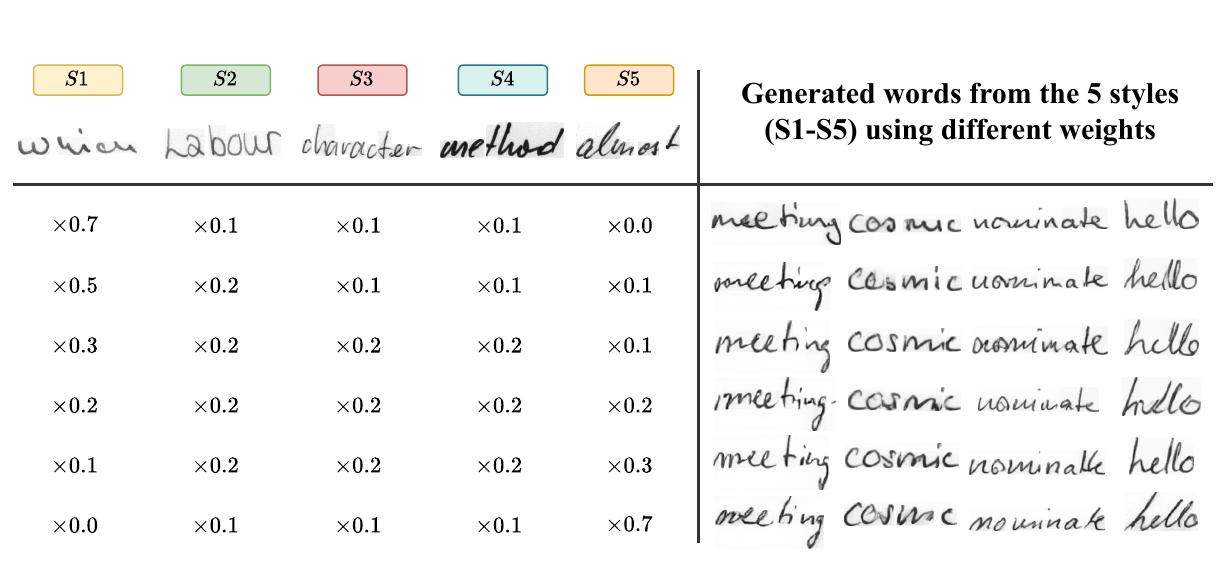}
        \caption{Style Mixture between 5 styles $S1-S5$.}
        \label{fig:style_mixture_supplementary1}
    \end{subfigure}
    \begin{subfigure}{0.9\textwidth}
        \centering
        \includegraphics[width=\linewidth]{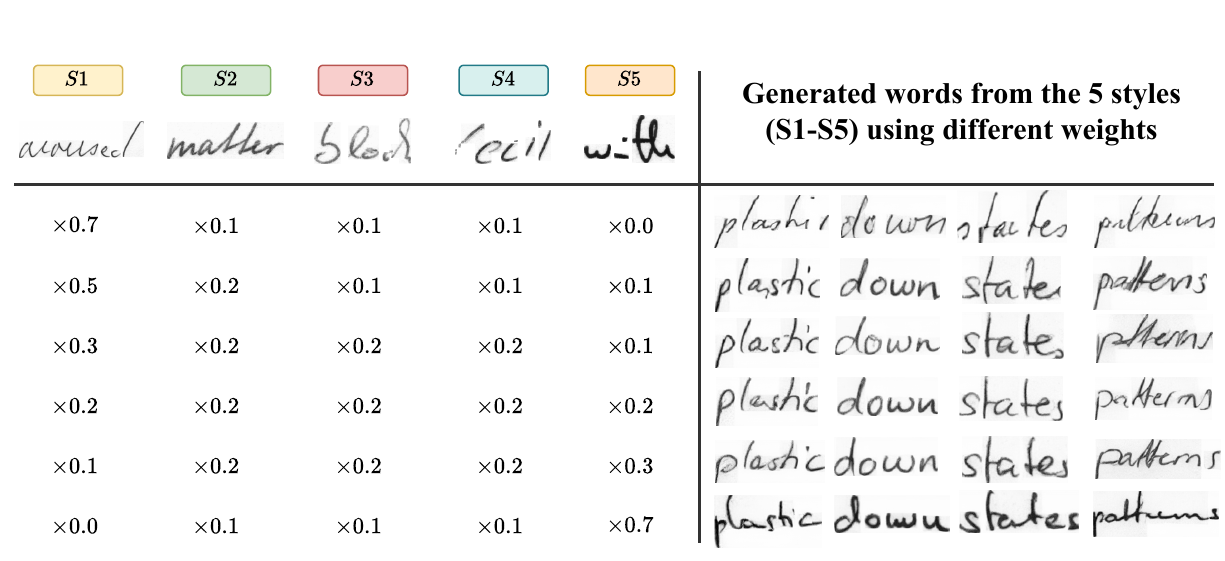}
        \caption{Style Mixture between 5 styles $S1-S5$.}
        \label{fig:style_mixture_supplementary2}
    \end{subfigure}
    \begin{subfigure}{0.9\textwidth}
        \centering
        \includegraphics[width=\linewidth]{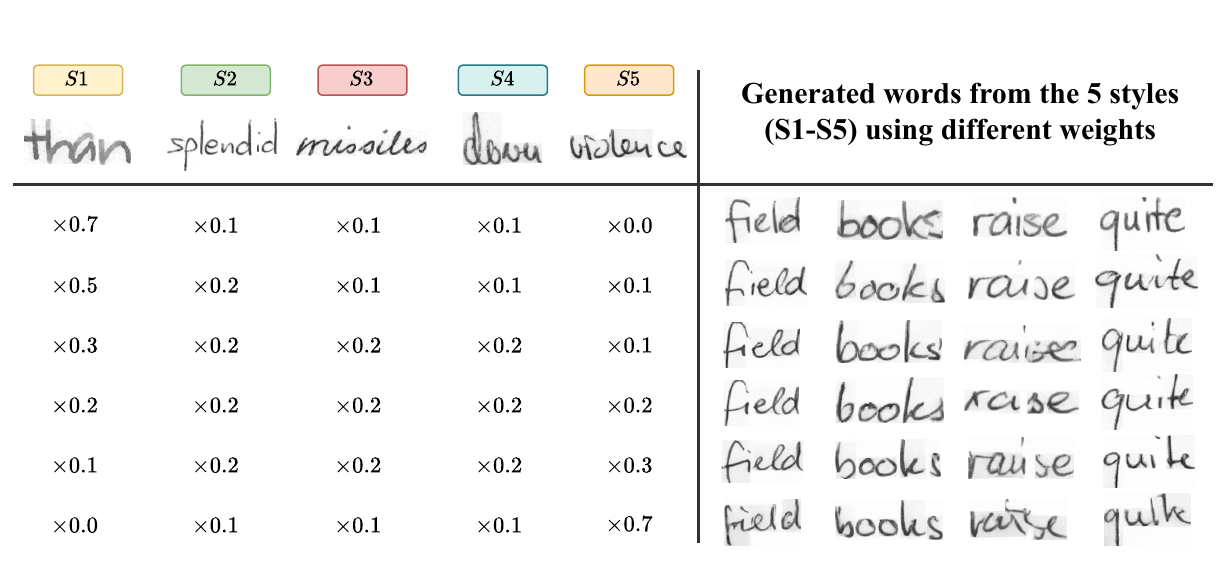}
        \caption{Style Mixture between 5 styles $S1-S5$.}
        \label{fig:style_mixture_supplementary3}
    \end{subfigure}
    
   \caption{Generated samples by combining five different writing styles.}
   \label{fig:style_mixture_supplementary}
\end{figure*}

\noindent
\textbf{Style Mixture.}
We extend the qualitative examples of Style Mixture presented in Figure 7 of the main paper and show the results in~\cref{fig:style_mixture_supplementary}. 
One can see that our method is able to generate new styles by mixing more than two and adjusting the weights.
This ability comes from modeling the style space and using the mean embedding of the 5 feature samples.
This is not the case for the comparing methods that perform the few-shot style condition, as they concatenate the style features, thus obtaining a different embedding every time.

\begin{figure}[!t]
\vspace{-5cm}
\centering
\includegraphics[width=\textwidth]{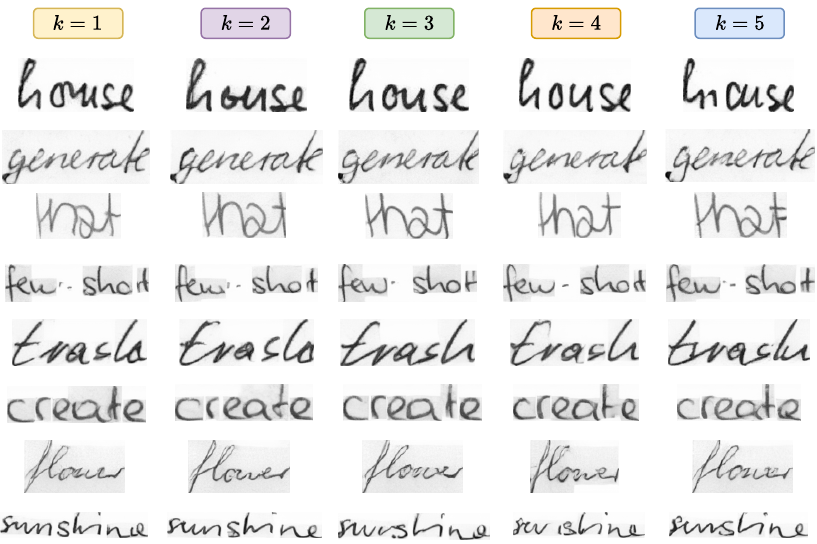}
    \caption{Effect of the number of style samples during sampling.}
    \label{fig:few_shot_sup}
\end{figure}

\noindent
\textbf{Few-Shot Style Effect.}
We explore how the number of style samples affects the generation during sampling.
We experiment with 1-5 samples to condition the word generation and present the results in~\cref{fig:few_shot_sup}.
One can see that although the model is trained with a condition of $k=5$ samples, it can still generate quality samples with fewer style images, even one.

\section{Handwriting Text Recognition}
\label{sec:htr_sup}
We present a more detailed analysis of the Handwriting Text Recognition (HTR) experiments present in our main paper using the CNN-LSTM HTR system presented in~\cite{retsinas2022best}, which is trained with Connectionist Temporal Classification (CTC) loss~\cite{graves2008novel}.
We train the HTR system using the real data of the IAM training set and explore the effect of additional synthetic sets.
~\cref{tab:htr_sup} shows the results of the generated data used as additional data to the real training set.
We observe that WordStylist~\cite{nikolaidou2023wordstylist} and our proposed method, DiffusionPen, show improved HTR performance in terms of Character Error Rate (CER) and Word Error Rate (WER), with WordStylist obtaining a slightly better performance.
It should be noted that the HTR results presented in the original paper of WordStylist~\cite{nikolaidou2023wordstylist} are only on a subset of IAM that excludes punctuation and words smaller than 2 characters and larger than 10 characters for both train and test set.
In our work, we have re-trained WordStylist~\cite{nikolaidou2023wordstylist}, GANwriting~\cite{kang2020ganwriting} and SmartPatch~\cite{mattick2021smartpatch}, using the full character set.
Thus, our obtained HTR performance for WordStylist differs from the one presented in the original paper~\cite{nikolaidou2023wordstylist}.

We further explore the effect of the style variation data as an augmentation of the existing training set.
We present results using the data generated using the noisy style embedding and the noise bias concepts mentioned in~\cref{sec:style_variation_sup} in~\cref{tab:htr_style_sup}.
We can see that both cases (Style Noise 0.25 and Noise Bias) can improve the performance of the HTR system; however, they cannot achieve a better performance than our standard method (see last row of~\cref{tab:htr_sup}).

\end{document}